\documentclass[]{fairmeta}
\usepackage{xspace}
\usepackage{wrapfig}
\usepackage{mathpazo}
\usepackage{tgpagella}
\usepackage{bbding}
\usepackage{graphicx}
\usepackage{pgfplots}
\pgfplotsset{compat=1.18}
\usepackage{makecell}
\usepackage{enumitem}
\usepackage{mathtools}
\usepackage{xfrac}
\usepackage{tikz}
\usepackage{colortbl}
\usetikzlibrary{spy}

\newcolumntype{L}[1]{>{\raggedright\let\newline\\\arraybackslash\hspace{0pt}}m{#1}}
\newcolumntype{C}[1]{>{\centering\let\newline\\\arraybackslash\hspace{0pt}}m{#1}}
\newcolumntype{R}[1]{>{\raggedleft\let\newline\\\arraybackslash\hspace{0pt}}m{#1}}
\newcolumntype{Y}{>{\centering\arraybackslash}X}

\title{EditCtrl: Disentangled Local and Global Control for Real-Time Generative Video Editing}
\author[1,2,\ast]{Yehonathan Litman}
\author[3]{Shikun Liu}
\author[1]{Dario Seyb}
\author[1]{Nicholas Milef}
\author[1]{Yang Zhou}
\author[1]{Carl Marshall}
\author[2]{\\Shubham Tulsiani}
\author[1]{Caleb Leak}

\affiliation[1]{Meta Reality Labs}
\affiliation[2]{Carnegie Mellon University}
\affiliation[3]{Meta AI}
\contribution[\ast]{Work done at Meta}

\abstract{
High-fidelity generative video editing has seen significant quality improvements by leveraging pre-trained video foundation models. However, their computational cost is a major bottleneck, as they are often designed to inefficiently process the full video context regardless of the inpainting mask's size, even for sparse, localized edits. In this paper, we introduce EditCtrl, an efficient video inpainting control framework that focuses computation only where it is needed. Our approach features a novel local video context module that operates solely on masked tokens, yielding a computational cost proportional to the edit size. This local-first generation is then guided by a lightweight temporal global context embedder that ensures video-wide context consistency with minimal overhead. 
Not only is EditCtrl $10\times$ more compute efficient than state-of-the-art generative editing methods, it even improves editing quality compared to methods designed with full-attention. 
Finally, we showcase how EditCtrl unlocks new capabilities, including multi-region editing with text prompts and autoregressive content propagation.

}

\date{\today}
\correspondence{\email{litman@cmu.edu}}
\metadata[Website]{\url{https://yehonathanlitman.github.io/edit_ctrl}}

\def\OurMethod{EditCtrl\xspace}
\def\frames{\mathbf{V}}
\definecolor{tabfirst}{rgb}{1, 0.7, 0.7} 
\definecolor{tabsecond}{rgb}{1, 0.85, 0.7} 
\definecolor{tabthird}{rgb}{1, 1, 0.7} 

\begin{document}

\maketitle

\begin{figure}[ht!]
    \centering
    \includegraphics[width=\textwidth, trim=0 4.4cm 0.2cm 0.2cm, clip]{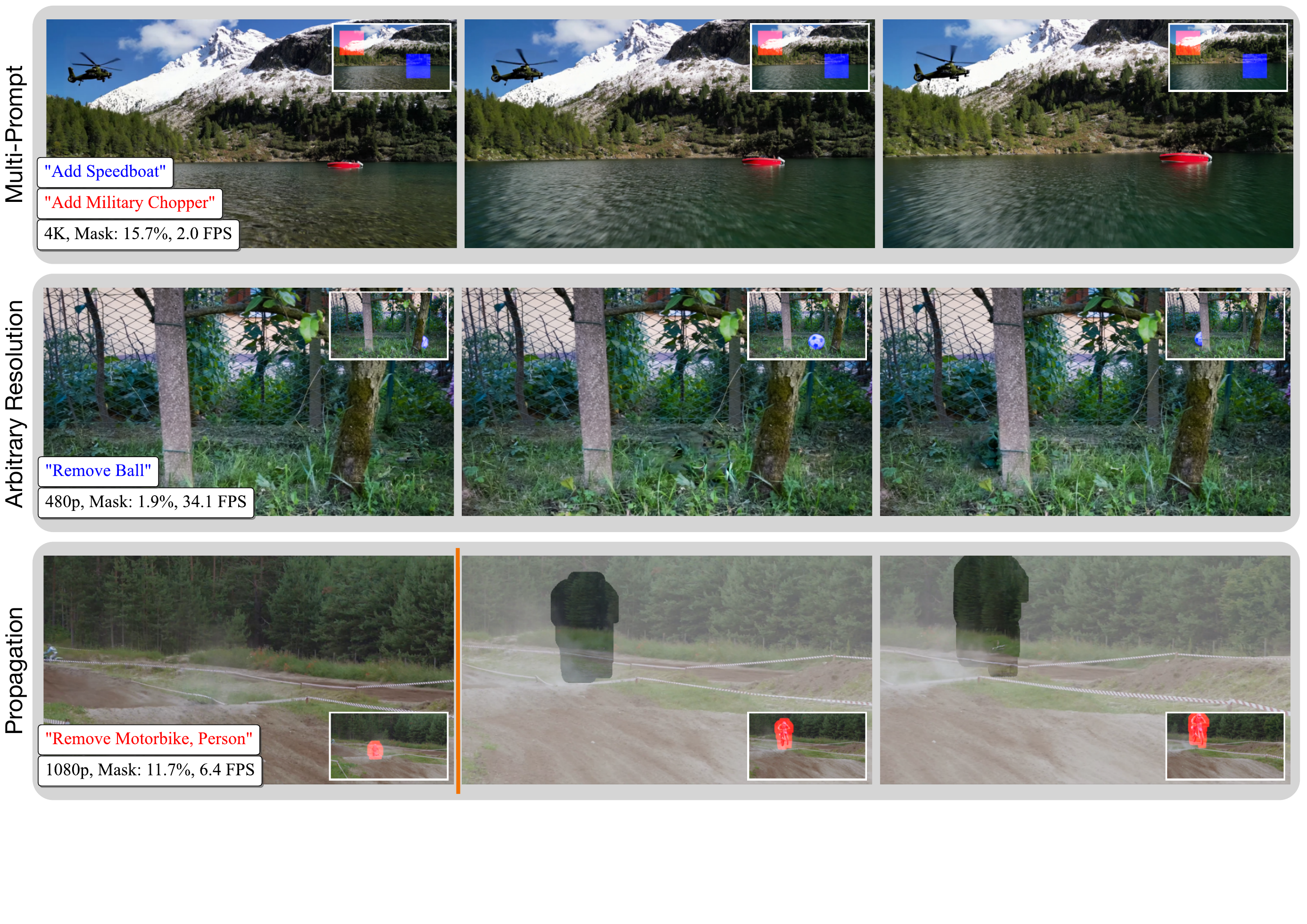}
    \vspace*{-0.4cm}
    \caption{\textbf{\OurMethod: A Real-time Generative Video Editing Pipeline.} \OurMethod supports complex, prompt-guided edits on 4K videos, simultaneously handling an arbitrary number of user-defined masks (Top). To maintain real-time performance, our inference pipeline dynamically allocates compute proportional to the edit mask size (Middle). \OurMethod also intelligently propagates object edits from initial frames into the future (after the orange line), ensuring high temporal and object consistency in the resulting edit (Bottom).}
    \label{fig:teaser}
\end{figure}

\section{Introduction}
\label{sec:introduction}

Generative video editing is a rapidly advancing field in media content creation, with potential applications ranging from professional film production to real-time augmented reality. A fundamental and challenging task within this domain is {\it video inpainting}: the process of replacing arbitrary regions of a video with high-fidelity, contextually consistent content. Although data-driven feed-forward approaches have shown success in simple content removal (e.g., object removal, background filling, deblurring) \citep{zhou2023propainter, Liu_2021_FuseFormer, E2FGVI, zhang2024blur, Zhang_2022_FGT}, they often struggle with generative tasks such as adding entirely new objects or large-scale scene replacements.

The advancement of large-scale text-to-video diffusion models \citep{wan2025, HaCohen2024LTXVideo, kong2024hunyuanvideo, polyak2024moviegen} has pushed the boundaries of generation and, in turn, enabled follow-up work on high-quality, semantically-aware video inpainting applications \citep{ma2025follow, huang2025vivid4d, zhang2023avid, liu2024generativevideopropagation, gen-omnimatte, yang2025matanyone}. However, video generation incurs significant computational costs, which makes current state-of-the-art inpainting methods \citep{bian2025videopainteranylengthvideoinpainting, vace, li2025diffueraserdiffusionmodelvideo} and applications inefficient. This is due to processing the entire spatiotemporal context of the video, regardless of whether an edit is needed in a given region. This dense, full-attention approach is profoundly inefficient, making it prohibitive for real-world applications that need fast inference, such as real-time augmented reality, editing high-resolution videos, or applying multiple distinct edits simultaneously.

We are inspired by prior work that overcomes this bottleneck by efficiently allocating computation for inpainting using tokens inside the target edit mask region \citep{zhou2023propainter, lazy_diffusion}. In particular, this principle can be applied in generative image editing by finetuning a pre-trained diffusion model to only focus on relevant local features while using a compressed global representation as context \citep{lazy_diffusion}. We extend this approach to tackle the challenges in generative video editing: understanding and inpainting temporally coherent content, and enabling intelligent video-specific downstream operations like content propagation. Specifically, we seek to leverage high-quality video generation models in a disentangled manner \emph{without} harming the pre-trained quality and achieve this via learning lightweight adapter modules.

In this paper, we propose \OurMethod, a disentangled generative video editing framework for temporally coherent video inpainting. Our design follows two principles: i) computation should scale with the edit region, not the video resolution; and ii) conditioning should be injected via adapters on a frozen base model, preserving its generative quality and compatibility with model variants. We instantiate these through a \textbf{sparse local context encoder} that processes only tokens within the edit mask, and a \textbf{lightweight temporal global context embedder} that provides video-wide cues to guide local generation.

Because \OurMethod achieves proportional compute scaling entirely through lightweight adapters on a frozen base model, it preserves the full generative quality of the pretrained backbone while grounding local generation in global temporal context for coherent edits. This non-destructive design makes \OurMethod{} {\it natively compatible} with model variants such as distilled and autoregressive models, unlocking downstream intelligent video editing applications, including multi-region editing with distinct prompts and real-time content propagation into future frames.

We demonstrate the effectiveness of our approach qualitatively and quantitatively, showing that it matches the performance of its full-attention base model and outperforms other state-of-the-art methods. Finally, we highlight the practical utility of our model by showcasing its applications in demanding scenarios, including fast arbitrary-resolution video editing, simultaneous multi-region inpainting, and real-time augmented reality content editing and propagation.

\section{Related Works}
\label{sec:related_works}

\paragraph{Generative Video Inpainting}
Early data-driven video inpainting methods used spatiotemporal transformers or convolutional networks to attend to the global context of the video while propagating information across frames using optical flow \citep{zhou2023propainter, E2FGVI, zhang2024blur, Zhang_2022_FGT}. These methods efficiently inpaint target regions, but are limited in their ability to handle complex motion, masked regions, or generate coherent semantic content given a condition like a text prompt. To increase video inpainting quality and diversity, generative-based methods repurposed powerful priors in the form of video diffusion models \citep{bian2025videopainteranylengthvideoinpainting, vace, li2025diffueraserdiffusionmodelvideo, liu2024generativevideopropagation, samuel2025omnimattezero, ma2025follow, gen-omnimatte}. While generative-based models inpaint high-quality content coherently, they take both the entire video as context and a per-pixel mask indicating whether to preserve or generate each pixel.
This approach is inefficient for interactive editing since the underlying diffusion model needs to repeatedly generate pixels outside the local masked edit area. Furthermore, it ties mask information to the video context, preventing generative inpainting from making multiple distinct edits at different regions or propagating inpainting information when video frames are not available in tasks like image-to-video generation.

\paragraph{Accelerating Video Generation}
To accelerate video editing, a straightforward approach is to reduce the number of diffusion sampling steps the base video diffusion model needs to produce coherent outputs. This includes knowledge distillation \citep{yin2024onestep, yin2025causvid, luo2023latent, heek2024multistepconsistencymodels}, sparse attention \citep{liu2025mardini, zhang2025vsa, shin2025motionstream, zhang2025spargeattn}, or linear transformers \citep{xie2024sana, chen2025sana}. This can be naively applied to the base video diffusion model used for generative editing, accelerating editing as a result. However, this only reduces the number of underlying diffusion steps and not the data processed for content editing in terms of the local and global context sizes. As such, runtime and memory scale with the input video resolution regardless of the target edit area. Token merging based approaches \citep{bolya2022tome, li2023vidtome, object_centric_video_diffsion} address this by pruning tokens outside the local context, though this leads to significant quality degradation and may actually slow inference due to the token importance calculation step. Some generation methods compress the global context together with the local context \citep{beyer2025highly}, but this requires training specialized generative models that can take the new information, which is expensive and not competitive with the quality of pretrained full-attention diffusion models. We instead adopt insights from an efficient image editing approach \citep{lazy_diffusion} that fine-tunes a pretrained diffusion model to only operate on local tokens with additional compressed global context, thus accelerating diffusion proportionally relative to the masked region area.

Our work extends this insight to the video domain while only learning local adapters instead of full finetuning, thus allowing integration with additional control variants without modifying the base model weights.

\paragraph{Controlled Video Generation}
As video editing is a task that requires spatiotemporal control, the local and global context can be considered as high-frequency and low-frequency control components, respectively. Recent work in controllable generation has shown success in guiding pretrained image and video models using external signals \citep{gu2025das, zhang2023controlvideo, zhang2023controlnet}. These signals include low-frequency ones, such as human pose parameters \citep{zhou2025realisdance-dit, hu2023animateanyone} and camera parameters \citep{wang2024motionctrl, he2025cameractrl, bahmani2024ac3d, bai2025recammaster}, or high-frequency ones such as segmentation and depth maps \citep{cai2023genren,nvidia2025cosmostransfer1conditionalworldgeneration, HaCohen2024LTXVideo, structuresynthesis} as well as environmental illumination and intrinsic decomposition \citep{litman2025lightswitch, liang2025luxdit, he2025unirelight, litman2025materialfusion}, allowing fine-grained control over generated content. In addition, overlaying multiple low and high-frequency control signals causes guidance to become a multimodal task \citep{nvidia2025cosmostransfer1conditionalworldgeneration, ruan2022mmdiffusion, corona2024vlogger}, but this has not been explored for accelerating generation given multiple control inputs.

\OurMethod uses lightweight adapters to disentangle the low and high frequency components of the mask, which correspond to the global and sparse local contexts, to significantly accelerate video editing with minimal interference in the base model. By keeping the base model frozen, our approach remains invariant to video resolution, enables fast editing, and remains compatible with additional control variants. It naturally extends to interactive video editing capabilities such as multi-prompt editing for different regions and temporal content propagation, which are challenging for non-disentangled frameworks that rely on full-attention.

\section{Methodology}
\label{sec:methodology}

\begin{figure*}
    \centering
    \includegraphics[width=\textwidth]{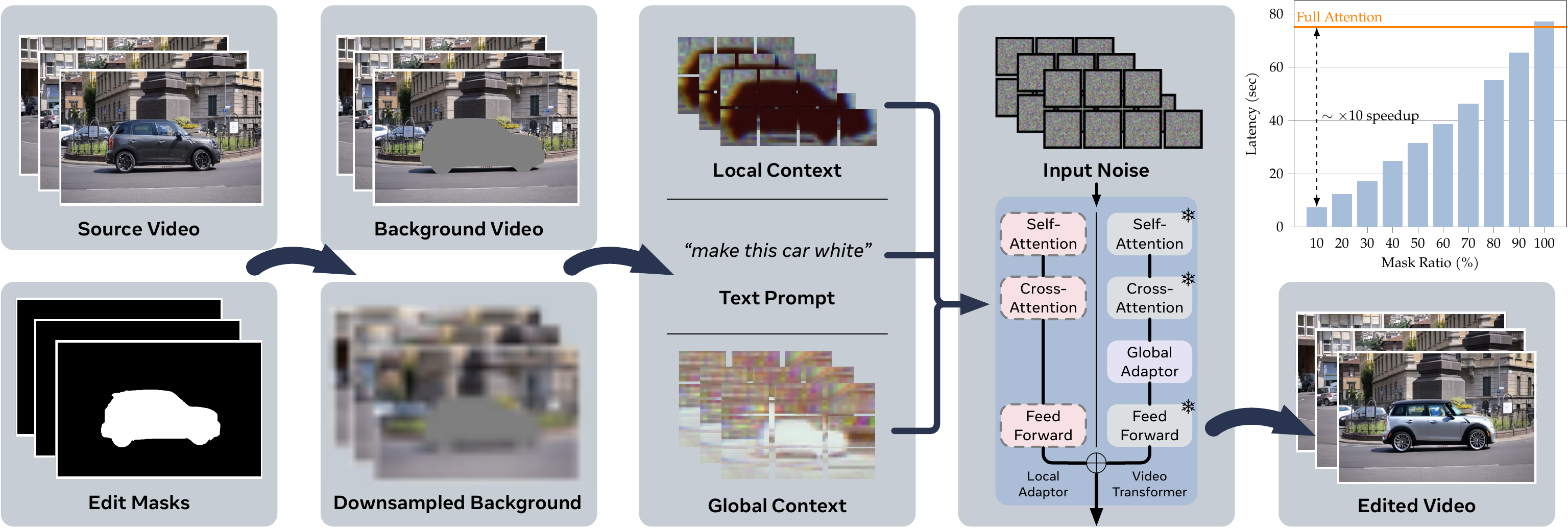}
    \caption{\textbf{\OurMethod Video Diffusion Framework Overview.} \OurMethod edits a source video given a target edit mask. Foreground content is masked out, giving the background video that is also down-sampled to a constant resolution regardless of the original resolution. The compact global context of the down-sampled background video and the local context at the mask edit region are then encoded. These are given to trainable local and global adapters inside a pretrained text-to-video diffusion model that \textit{denoises tokens $\mathbf{z}^t$ only in the masked edit region} given a text prompt. After diffusion, the tokens are scattered into the masked edit region in the encoded source video latents. Our method shows a proportional speedup with respect to the target mask area ratio.}
    \label{fig:editctrl_overview}
\end{figure*}

In this section, we introduce our video editing framework {\bf \OurMethod} (Fig.~\ref{fig:editctrl_overview}), which accepts an input video, a corresponding masked video, and a text prompt to produce the edited video with a computational saving proportional to the number of unmasked pixels. We first outline our notation and review the preliminaries of video diffusion (Sec.~\ref{sec:preliminaries}). We then describe the framework's foundation and how it is derived from a base inpainting control module (Sec.~\ref{sec:framework_overview}). Finally, we show how our approach can be adapted at inference time for interactive tasks such as multi-region inpainting and also real-time video editing and propagation for augmented reality (Sec.~\ref{sec:edit_capabilities}).

\begin{figure*}
    \centering
\includegraphics[width=\textwidth]{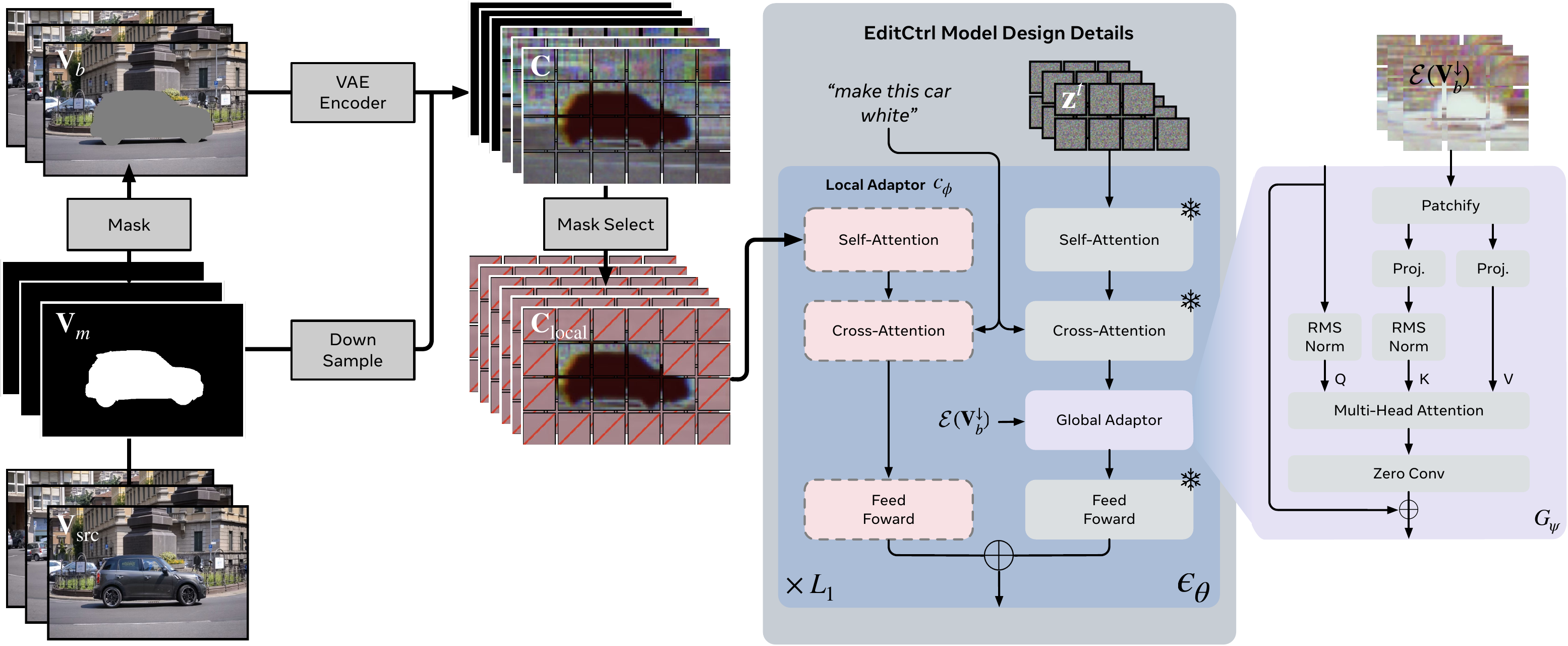}
    \caption{\textbf{\OurMethod: Local and Global Control Modules.} Given the source video $\frames_\text{src}$ and target edit masks $\frames_m$, we extract the background content $\frames_b$ and encode it with a video VAE encoder $\mathcal{E}$. This is then concatenated channel-wise with the down-sampled masks to give the control context $\mathbf{C}$. Tokens in $\mathbf{C}$ outside the down-sampled edit mask region are then masked out, giving the local context tokens $\mathbf{C}_\text{local}$ which go to the local encoder module $c_\phi$, whose outputs are added to selected transformer layers. The global embedder $G_\psi$ receives the query feature tokens and global context tokens produced from the down-sampled background content $\frames_b^\downarrow$ and modulates the noisy cross-attended features.}
    \label{fig:local_encoder_global_embedder}
\end{figure*}

\subsection{Preliminaries and Notation}
\label{sec:preliminaries}

\paragraph{Text-to-Video Diffusion.}
Text-to-video diffusion models are primarily structured as latent diffusion models using a powerful diffusion transformer (DiT) backbone $\epsilon_\theta$ to generate visually appealing, temporally consistent content given a text prompt $\mathbf{p}$. The model employs a pretrained variational autoencoder (VAE) consisting of an encoder $\mathcal{E}$ and decoder $\mathcal{D}$ to encode videos to a smaller latent space and reduce computational cost. 
At training, $\mathcal{E}$ encodes a source video $\frames_\text{src}\in \mathbb{R}^{F \times H\times W}$, defined as a sequence of $F$ frames of size $H\times W$, into latent space: $\mathbf{z}=\mathcal{E}(\frames_\text{src})$. A sampled noise $\epsilon_t$ is added to the encoded latent using a scheduler \citep{ho2020ddpm} with timestep $t\in\{1, \cdots, T\}$. $\epsilon_\theta$ takes the produced noisy video latent $\mathbf{\bar{z}}^t$ to predict the noise $\epsilon_t$ in the forward diffusion process:
\begin{align}
    \mathcal{L}_{\text{DM}} = 
    \| \epsilon_\theta\left(\mathbf{\bar{z}}^t, t; \mathbf{p} \right) - \epsilon_t \|_2^2,
\end{align}

\paragraph{Control Adapters.} 
To make use of the generative prior knowledge in pretrained text-to-video diffusion models for video editing, a conditional adapter module $c_\phi$ is trained to compress additional conditional information $\mathbf{C}$ and steer the predictions from $\epsilon_\theta$. $\mathbf{C}$ is derived from editing mask frames $\frames_m\in \mathbb{R}^{F \times H\times W}$ corresponding to target edit regions. These are used to mask the foreground in the input video, giving the background video $\frames_b\in \mathbb{R}^{F \times H\times W}$ that is encoded with $\mathcal{E}$. $\mathbf{C}$ is constructed as a channel-wise concatenation of the information $(\mathcal{E}(\frames_b), \frames_m^\downarrow)$, where ${\frames_m^\downarrow}$ are the editing masks that are down-sampled to the latent resolution, which gives our conditional diffusion process:
\begin{equation}
    \mathcal{L}_{\text{CDM}} = 
    \| \epsilon_\theta\left(\mathbf{\bar{z}}^t, t; \mathbf{p}, c_\phi(\mathbf{C}) \right) - \epsilon_t \|_2^2,
\end{equation}

\paragraph{Context-Guided Local Computation.}
LazyDiffusion \citep{lazy_diffusion} introduced a principle for diffusion-based efficient (image) editing where computation is restricted to the edit region and edit context can be supplied through a compact side channel. A diffusion model is fine-tuned to process noisy latent tokens $\mathbf{z}^t$ strictly within the masked region while a separate encoder captures local and global information context $\mathbf{C}_\text{enc}$. $\mathbf{C}_\text{enc}$ is concatenated to the hidden dimension at every denoising iteration, achieving acceleration proportional to the mask size with global context. A pre-trained diffusion model is finetuned with a ``lazy'' diffusion process:
\begin{align}
    \mathcal{L}_{\text{Lazy-DM}} = 
    \| \epsilon_\theta\left(\mathbf{z}^t\oplus \mathbf{C}_\text{enc}, t; \mathbf{p} \right) - \epsilon_t \odot \frames_m^\downarrow \|_2^2,
\end{align}

\subsection{Disentangled Editing Architecture Design}
\label{sec:framework_overview}

Our goal is to design an efficient video editing framework that scales proportionally with respect to edit area and flexibly composes with adapters that were pre-trained for full-attention tasks. Towards this, we build on the design choices highlighted in Sec.~\ref{sec:preliminaries} of achieving flexibility with control adapters and efficiency via context-guided local computation. Specifically, we i) disentangle video editing with local encoder and global embedder adapters that separately operate on local context tokens in addition to efficiently encoding the full global temporal context to maintain coherence, and ii) achieve this with adapter-based finetuning. This design combination focuses compute in masked regions with temporal coherence, preserves the base model's generative capabilities, and enables intelligent downstream video editing applications. 

\paragraph{Context-Guided Video Diffusion}
\OurMethod builds on a ControlNet-like architecture \citep{zhang2023controlnet}, with a trainable context control module used to guide a frozen, pre-trained video diffusion model \citep{HaCohen2024LTXVideo, vace, nvidia2025cosmostransfer1conditionalworldgeneration}. This context module is trained for full-frame inpainting with bi-directional full-attention, processing the entire video $\frames_\text{src}$ and editing masks $\frames_m$, making it computationally expensive regardless of the edit size. To repurpose the inpainting control module for local context editing, we use $\frames_m^\downarrow$ as an attention mask for the tokens corresponding to the foreground. Background tokens $\mathbf{C}$ are selected with $\frames_m^\downarrow$ to produce $\mathbf{C}_\text{local}$. 
Masking out all the background tokens results in poor blending, so $\frames_m^\downarrow$ is dilated to neighboring pixels before selection \citep{Liu_2021_FuseFormer}. 
This attention mask operation is also applied to $\mathbf{z}^t$, and the output from the control module is summed element-wise with the feed-forward network output after selected video transformer layers.
By processing only the corresponding local tokens through the video diffusion model and the control module, the diffusion process is proportionally accelerated. However, generation quality is much worse as the pretrained control module was trained with full-attention and not sparse attention patterns corresponding to the target mask region. Thus, the module is fine-tuned into a local context encoder using a mask-aware diffusion loss
\begin{align}
    \mathcal{L}_\phi = 
    \| \epsilon_\theta\left(\mathbf{z}^t, t; \mathbf{p}, c_\phi(\mathbf{C}_\text{local}) \right) - \epsilon_t \odot \frames_m^\downarrow \|_2^2,
\end{align}

The fine-tuned local context encoder enables high-quality local edit generation without access to the full spatiotemporal context. At inference, $\mathbf{z}^0$ is scattered into $\mathcal{E}(\mathbf{V}_\text{src})$ for the final edited video output.

\paragraph{Cross Attention Modulation.} To incorporate global information such as appearance and scene cues (e.g., lighting, structure, dynamics, camera motion) at generation, we enrich the cross-attended features $\mathbf{x}$ produced with the video DiT by injecting temporally-aware attention features as shown in Fig.~\ref{fig:local_encoder_global_embedder}. This is similar to using CLIP for image-guided generation, but in our case we use a temporal global context representation computed using the VAE encoder $\mathcal{E}(\frames_b^\downarrow)$, denoted as $\mathbf{C}_\text{global}$. $\frames_b$ is spatially down-sampled to a resolution of $256\times 256$ regardless of the original resolution to give $\frames_b^\downarrow$. This is to increase robustness to aspect ratio and the number of frames we can encode. $\mathbf{C}_\text{global}$ is then processed by a trainable patch layer to produce the global context token embeddings, efficiently capturing video-wide temporal evolution and high level scene cues while remaining invariant to the original video resolution. The global embedder $G_\psi$ takes the global context embeddings to calculate the attention weights of query tokens $Q$ with the global feature keys and values $\mathbf{K}_g, \mathbf{V}_g$. These are added to the features after cross-attending to the text prompt
\begin{align}
    \mathbf{x} = \mathbf{x} + \mathbf{W}_0\cdot \text{Attention}(\mathbf{Q}, \mathbf{K}_g,\mathbf{V}_g)
\end{align}

where $\mathbf{W}_0$ is a zero initialized linear layer. This attention operation enables minimal yet sufficient control over the text prompt embedding, enriching the features without degrading the information from the text embedding. $G_\psi$ thus steers the generation of local token features to ensure they align with the text prompt and high level global video context with minimal compute overhead. We define the mask-aware loss function with $\mathbf{C}_\text{global}$ as
\begin{align}
    \mathcal{L}_\psi = 
    \| \epsilon_\theta\left(\mathbf{z}^t, t; \mathbf{p}, G_\psi(\mathbf{C}_\text{global}), c_\phi(\mathbf{C}_\text{local}) \right) - \epsilon_t \odot \frames_m^\downarrow \|_2^2,
\end{align}

Using both $\mathcal{L}_\phi$ and $\mathcal{L}_\psi$ from the start makes training unstable, however, as $c_\phi$ learns to generate local content given the text prompt and $G_\psi$ complicates that by modifying cross-attended features. Conversely, $G_\psi$ is not able to better guide the generation since $c_\phi$ gives poor predictions at the beginning of training. We use a piecewise training loss function $\mathcal{L}$ defined as a combination of $\mathcal{L}_\phi$ and $\mathcal{L}_\psi$
\begin{align}
    \mathcal{L} =
    \begin{cases}
    \mathcal{L}_\phi, & \text{if } k < n, \\[6pt]
    \mathcal{L}_\psi, & \text{if } k \geq n.
    \end{cases}
    \label{eq:total_loss}
\end{align}
where $k$ is the training iteration and $n$ is the predefined number of training iterations.

\subsection{Interactive Editing}
\label{sec:edit_capabilities}

The disentangled nature of \OurMethod enables efficient, flexible video editing that scales with the masked region rather than full video resolution. This enables several interactive applications, shown in Fig.~\ref{fig:teaser}, which full-attention methods do not naturally lend themselves to. First, as computational cost is independent of the overall video size, \OurMethod supports efficient editing of very high-resolution videos. Moreover, editing content only within the target area allows the framework to handle multiple, distinct edits simultaneously by batch-processing individual regions and merging the results into their respective locations in the output latents. This is not possible with a full-attention control module that uses the full binary video mask. Finally, our approach also facilitates real-time content propagation where future frames are unavailable. \OurMethod can be deployed together with an autoregressive video diffusion model \citep{huang2025selfforcing} to propagate edited content into the future and place it in frames when they are acquired. The high frame rate of videos means the global context does not change much, and simply padding the initial frames of $\frames_b^\downarrow$ with themselves provides sufficient global context, in addition to using the last frame's propagated background for local context. This enhances real-time video editing and eliminates latency for applications such as augmented reality, since we are generating and propagating into the future. Additional details and results are provided in the Appendix.

\begin{figure*}[t]
    \centering
    \begin{subfigure}[t]{0.5065\textwidth}
        \centering
        \includegraphics[width=\textwidth, trim=0.26cm 0pt 0 0pt, clip]{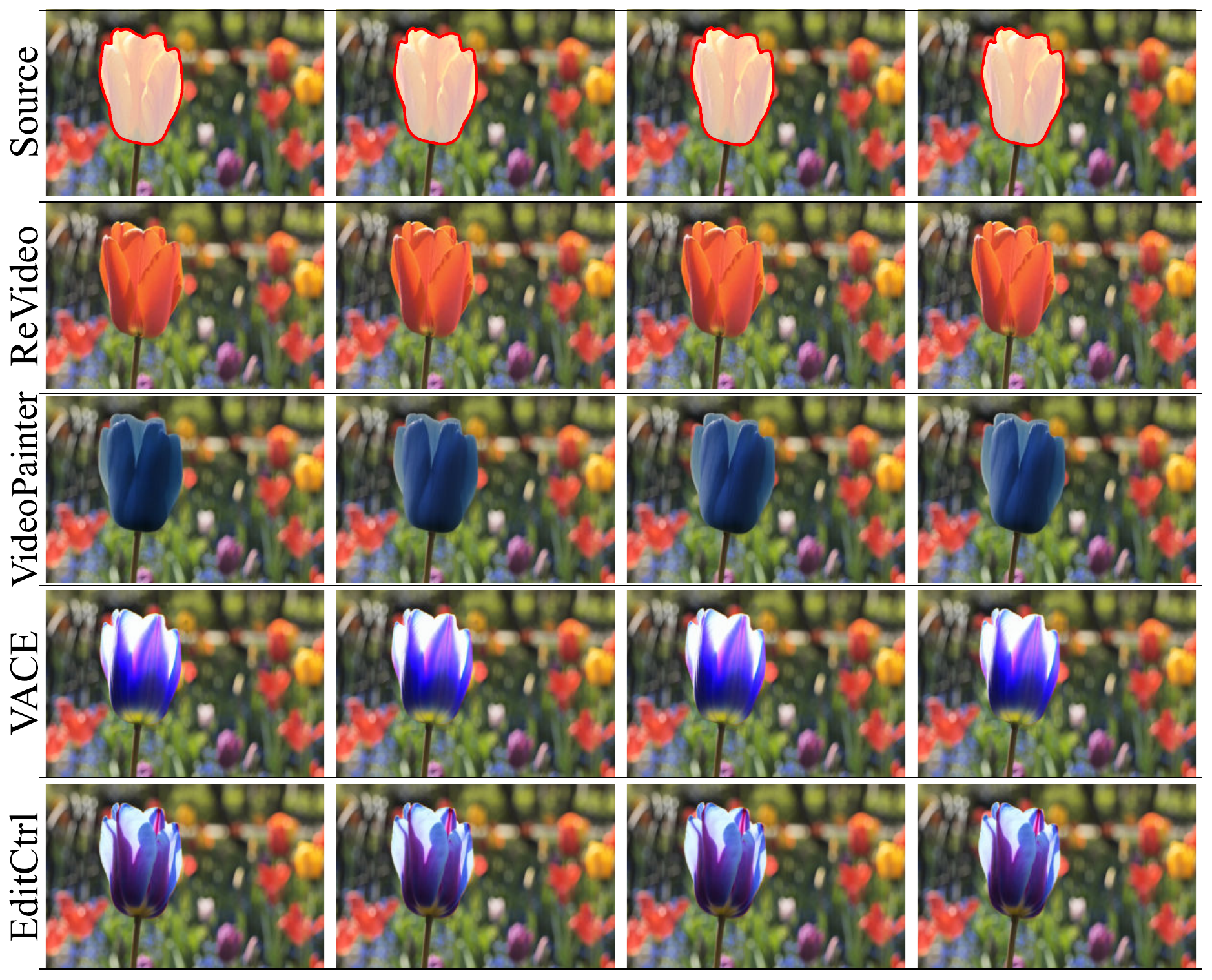}
        \vspace*{-0.45cm}
        \caption{``Change the flower color to blue.''}
    \end{subfigure}
    \hspace*{-0.2cm}
    \begin{subfigure}[t]{0.494\textwidth}
        \centering
        \includegraphics[width=\textwidth, trim=0pt 0pt 0.26cm 0pt, clip]{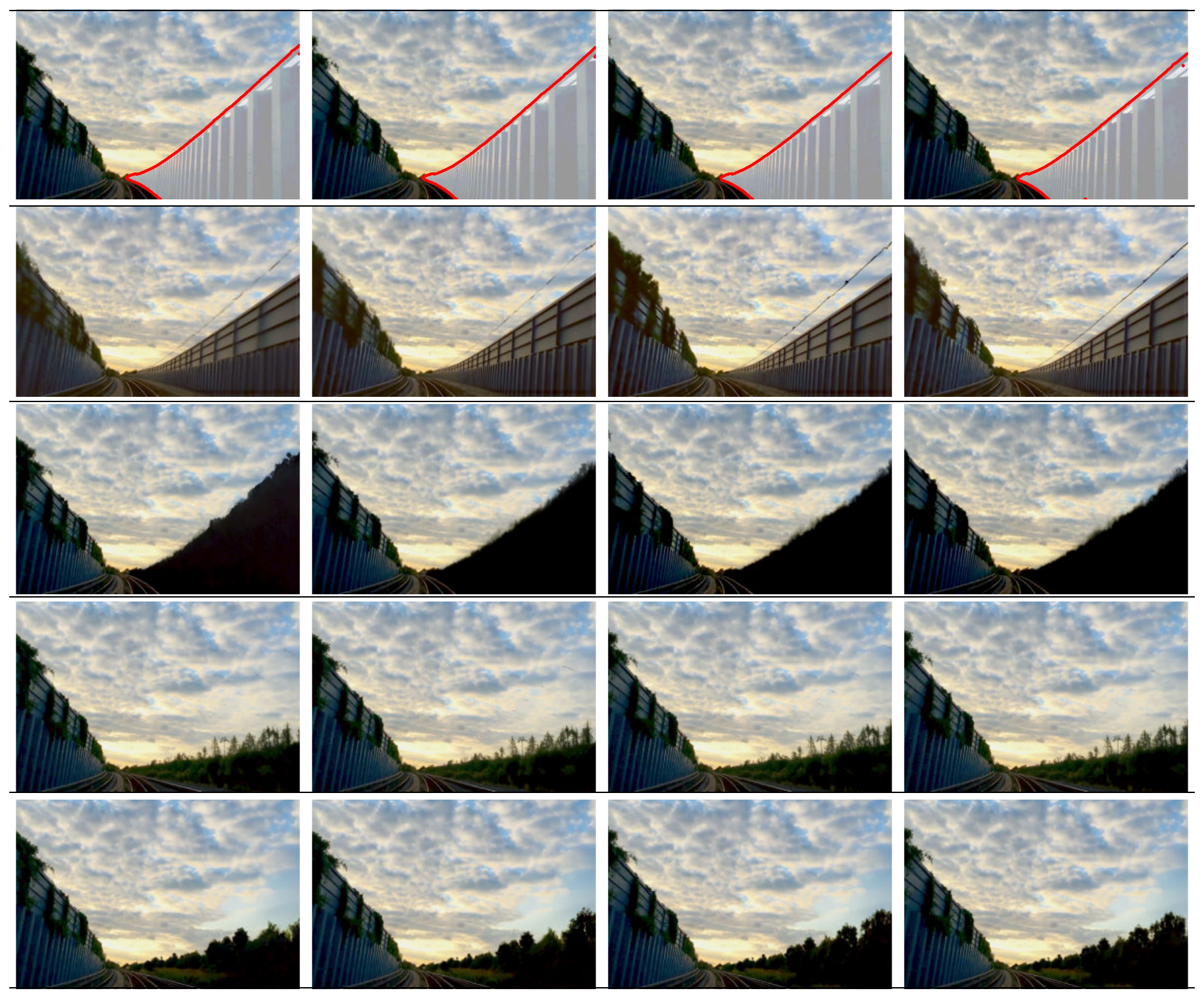}
        \vspace*{-0.45cm}
        \caption{``Remove the railing.''}
    \end{subfigure}

    \caption{\textbf{Video Editing Comparison.} \OurMethod generates visually appealing and structurally coherent edited content while the baselines either fail to edit the video correctly or produce content with poor appearance and blending. \OurMethod's localized editing greatly increases efficiency and enables real-time generative editing.}
    \label{fig:editing_visualization}
\end{figure*}

\begin{table*}[t]
\centering
\small
\setlength{\tabcolsep}{0.60mm}
\resizebox{\linewidth}{!}{
\begin{tabular}{lccccccc|cc|c|c}
\toprule 
\multirow{2}[3]{*}{\textbf{Method}} & \multirow{2}[3]{*}{\textbf{\#Par.}} & \multirow{2}[3]{*}{\textbf{PFLOPS$\downarrow$}} & \multicolumn{5}{c}{\bfseries Masked Region Preservation} 
& \multicolumn{2}{c}{\bfseries Text Alignment} 
& \multicolumn{1}{c}{\bfseries Temp. Coherence} 
& \multicolumn{1}{c}{\bfseries Throughput} \\
\cmidrule(lr){4-8} \cmidrule(lr){9-10} \cmidrule(lr){11-11} \cmidrule(lr){12-12}
& & & PSNR$\uparrow$ & SSIM$\uparrow$ & LPIPS$_{^{\times 10^2}}$$\downarrow$ 
& MSE$_{^{\times 10^2}}$$\downarrow$ & MAE$_{^{\times 10^2}}$$\downarrow$ 
& CLIP$\uparrow$ & CLIP (M)$\uparrow$ 
& CLIP Sim$\downarrow$ & FPS$\uparrow$ \\
\midrule
ReVideo \citep{mou2024revideo} & 1.5B & 193.39 & 15.52 & 0.49 & 27.68 & 3.49 & 11.14 & 9.34 & 20.01 & 0.42 & 0.11 \\  
VideoPainter \citep{bian2025videopainteranylengthvideoinpainting} & 5B & 817.81 & 22.63 & \cellcolor{tabthird}0.91 & 7.65 & \cellcolor{tabthird}1.02 & 2.90 & 8.67 & 20.20 & \cellcolor{tabthird}0.18 & 0.12 \\ 
VACE \citep{vace} & 1.3B & \cellcolor{tabsecond}76.31 & 23.84 & \cellcolor{tabthird}0.91 & \cellcolor{tabthird}5.44 & 0.92 & \cellcolor{tabsecond}2.78 & \cellcolor{tabsecond}9.76 & 21.51 & \cellcolor{tabfirst}0.13 & \cellcolor{tabthird}0.66 \\ 
VACE \citep{vace} & 14B & 589.19 & \cellcolor{tabthird}24.02 & \cellcolor{tabsecond}0.92 & \cellcolor{tabsecond}5.13 & \cellcolor{tabsecond}0.84 & \cellcolor{tabthird}2.68 & \cellcolor{tabfirst}9.85 & \cellcolor{tabthird}21.54 & \cellcolor{tabfirst}0.13 & 0.10 \\ 
 \midrule
 \OurMethod & 1.5B & \cellcolor{tabfirst}17.42 & \cellcolor{tabsecond}24.16 & \cellcolor{tabsecond}0.92 & 5.54 & 0.99 & 3.01 & \cellcolor{tabthird}9.58 & \cellcolor{tabsecond}21.70 & \cellcolor{tabsecond}0.15 & \cellcolor{tabfirst}4.67 \\ 
 \OurMethod & 16B & \cellcolor{tabthird}124.53 & \cellcolor{tabfirst}24.37 & \cellcolor{tabfirst}0.93 & \cellcolor{tabfirst}5.10 & \cellcolor{tabfirst}0.80 & \cellcolor{tabfirst}2.65 & 9.46 & \cellcolor{tabfirst}21.73 & \cellcolor{tabfirst}0.13 & \cellcolor{tabsecond}1.19 \\ 
\bottomrule
\end{tabular}
}
\caption{\textbf{Video Editing Comparison on VPBench-Edit.} \OurMethod outperforms editing baselines as well as the full attention base model in edited video quality, background preservation, and content alignment with the text prompt, all at a much higher throughput and lower computational cost (PFLOPS). In each column, the \colorbox{tabfirst}{best}, \colorbox{tabsecond}{second best}, and \colorbox{tabthird}{third best} results are marked.}
\vspace*{-0.3cm}
\label{tab:edit}
\end{table*}

\section{Experiments}
\label{sec:experiments}
\begin{figure*}
    \centering
    \captionsetup{type=figure}
        \includegraphics[width=0.5065\textwidth, trim=0.26cm 0pt 0 0pt, clip]{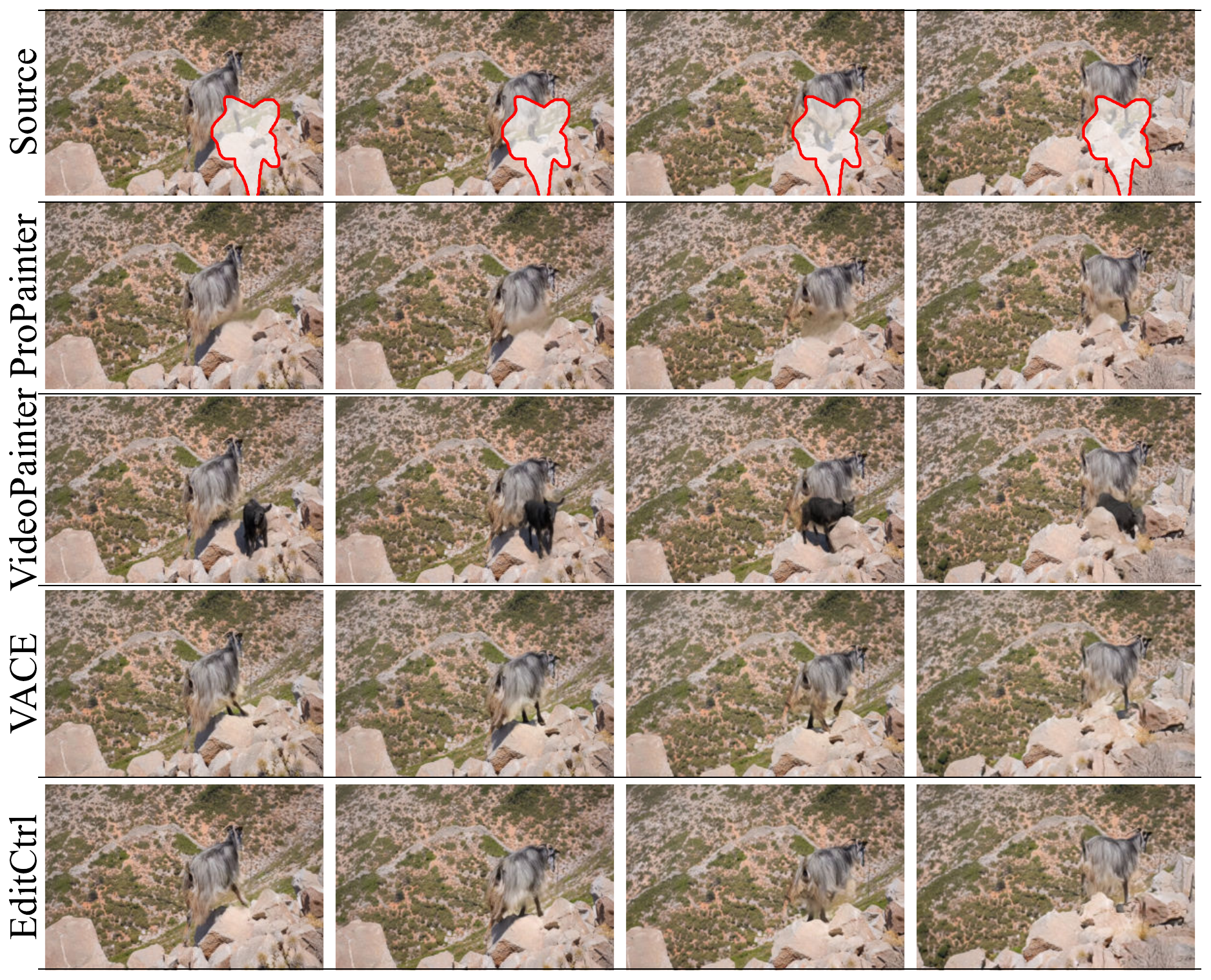}
        \hspace*{-0.2cm}
        \includegraphics[width=0.494\textwidth, trim=0pt 0pt 0.26cm 0pt, clip]{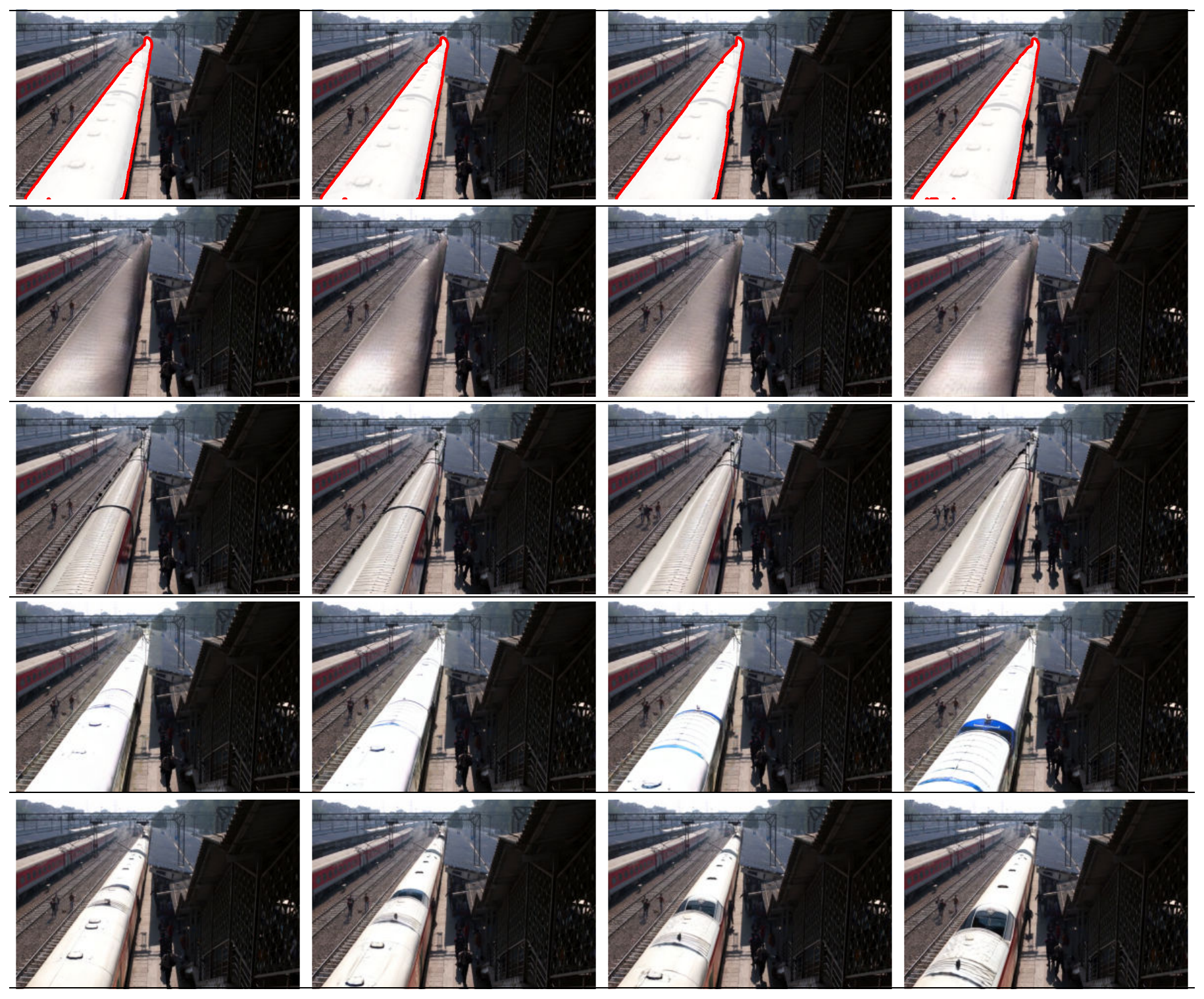}
       \captionof{figure}{
            \textbf{Video Inpainting Comparison.} Even with full-attention, baseline methods struggle to inpaint content that is coherent and visually appealing, while our method successfully generates high fidelity content that aligns with the scene using much less compute.
        }
        \label{fig:inpainting_visualization}
\end{figure*}

We evaluate on diverse video datasets labeled with target edit masks and detailed prompts to showcase \OurMethod's performance and efficiency in video editing tasks. We compare against previous video editing and inpainting baselines. We showcase \OurMethod's editing and inpainting performance, compute efficiency, and generalization with complex prompts, masks, and video content. We further provide ablations on our method to demonstrate how the local encoder and global embedder contribute to generative editing quality while reducing compute.

\subsection{Experimental Setup}

\begin{table*}[t]
\centering
\small
\setlength{\tabcolsep}{0.61mm}
\resizebox{\linewidth}{!}{%
\begin{tabular}{clcccccc|cc|c|c}
\toprule
\multicolumn{2}{c}{} & \multicolumn{1}{c}{} & \multicolumn{5}{c}{\bfseries Masked Region Preservation} & \multicolumn{2}{c}{\bfseries Text Alignment} & \multicolumn{1}{c}{\bfseries Temp. Coherence} & \multicolumn{1}{c}{\bfseries Throughput} \\
\cmidrule(lr){4-8} \cmidrule(lr){9-10} \cmidrule(lr){11-11} \cmidrule(lr){12-12}
 & \bfseries Method & \bfseries \#Par.
& PSNR$\uparrow$ & SSIM$\uparrow$ & LPIPS$_{^{\times 10^2}}$$\downarrow$ & MSE$_{^{\times 10^2}}$$\downarrow$ & MAE$_{^{\times 10^2}}$$\downarrow$ 
& CLIP$\uparrow$ & CLIP (M)$\uparrow$ & CLIP Sim$\downarrow$ & FPS$\uparrow$ \\
\midrule

& ProPainter \citep{zhou2023propainter} & 50M & 20.97 & \cellcolor{tabthird}0.87 & 9.89 & 1.24 & 3.56 & 7.31 & 17.18 & 0.44 & \cellcolor{tabfirst}5.34 \\
& VideoPainter \citep{bian2025videopainteranylengthvideoinpainting} & 5B & \cellcolor{tabsecond}23.32 & \cellcolor{tabfirst}0.89 & \cellcolor{tabfirst}6.85 & \cellcolor{tabfirst}0.82 & \cellcolor{tabfirst}2.62 & 8.66 & 21.49 & \cellcolor{tabsecond}0.15 & 0.12 \\
& VACE \citep{vace} & 1.3B & 22.62 & 0.85 & 9.30 & 1.25 & 4.43 & \cellcolor{tabsecond}9.77 & 21.55 & \cellcolor{tabthird}0.17 & 0.66 \\
& VACE \citep{vace} & 14B & 23.03 & \cellcolor{tabsecond}0.88 & \cellcolor{tabsecond}7.65 & \cellcolor{tabsecond}0.98 & \cellcolor{tabsecond}3.40 & 9.27 & \cellcolor{tabfirst}22.18 & \cellcolor{tabfirst}0.14 & 0.10 \\
  \midrule
& \OurMethod  & 1.3B & \cellcolor{tabthird}23.17 & 0.86 & 8.52 & 1.29 & 3.91 & \cellcolor{tabfirst}9.81 & \cellcolor{tabthird}21.86 & \cellcolor{tabthird}0.17 & \cellcolor{tabsecond}5.24 \\
& \OurMethod  & 14B  & \cellcolor{tabfirst}23.60 & \cellcolor{tabsecond}0.88 & \cellcolor{tabthird}8.23 & \cellcolor{tabthird}1.11 & \cellcolor{tabthird}3.63 & \cellcolor{tabthird}9.58 & \cellcolor{tabsecond}21.96 & \cellcolor{tabsecond}0.15 & \cellcolor{tabthird}1.30 \\
\midrule 
\midrule

& ProPainter \citep{zhou2023propainter} & 50M & 23.99 & 0.92 & 5.86 & 0.98 & 2.48 & \cellcolor{tabthird}7.54 & 16.69 & 0.12 & \cellcolor{tabsecond}5.51 \\
& VideoPainter \citep{bian2025videopainteranylengthvideoinpainting} & 5B & 25.27 & \cellcolor{tabfirst}0.94 & \cellcolor{tabfirst}4.29 & 0.45 & \cellcolor{tabfirst}1.41 & 7.21 & \cellcolor{tabthird}18.46 & \cellcolor{tabfirst}0.09 & 0.12 \\
& VACE \citep{vace} & 1.3B & \cellcolor{tabthird}25.75 & 0.88 & \cellcolor{tabthird}5.11 & \cellcolor{tabsecond}0.34 & 2.03 & 7.27 & 17.83 & 0.10 & 0.66 \\
& VACE \citep{vace} & 14B & \cellcolor{tabfirst}26.12 & \cellcolor{tabsecond}0.91 & \cellcolor{tabsecond}4.88 & \cellcolor{tabfirst}0.33 & \cellcolor{tabthird}2.01 & \cellcolor{tabfirst}7.81 & \cellcolor{tabfirst}18.75 & \cellcolor{tabfirst}0.09 & 0.10 \\
  \midrule
& \OurMethod & 1.5B & 25.44 & 0.86 & 5.31 & \cellcolor{tabthird}0.39 & \cellcolor{tabthird}2.01 & 7.33 & 18.02 & 0.12 & \cellcolor{tabfirst}5.57 \\
& \OurMethod & 16B & \cellcolor{tabsecond}25.89 & \cellcolor{tabthird}0.90 & 5.25 & \cellcolor{tabsecond}0.34 & \cellcolor{tabsecond}1.99 & \cellcolor{tabsecond}7.78 & \cellcolor{tabsecond}18.50 & \cellcolor{tabsecond}0.10 & \cellcolor{tabthird}1.41 \\
\bottomrule
\end{tabular}
}
\caption{\textbf{Video Inpainting Comparison.} \OurMethod matches or outperforms inpainting baselines for the VPBench-Inp (top) and DAVIS (bottom) test sets while significantly increasing inpainting efficiency.}
\label{tab:inp}
\end{table*}

\paragraph{Datasets.}
For training the local encoder and global embedder adapters, we used an internal dataset of diverse stock footage videos with segmentation annotations and descriptive text captions. Similar to VACE \citep{vace}, \OurMethod was trained on videos annotated with text captions, and the edit instructions in the figures are abstractions of translated target captions. We evaluate \OurMethod on the VPBench-Edit test set and the DAVIS and VPBench-Inp video inpainting test sets \citep{bian2025videopainteranylengthvideoinpainting, davis_2017}. The masks are randomly augmented during training \citep{suvorov2021resolution} to enable editing with diverse unstructured edit masks. This allows for inpainting on randomly shaped and placed masks, such as those in DAVIS, and structured segmentation maps found in VPBench. Both test datasets are publicly available. 

\paragraph{Fine-tuning Details.}
We initialize the local encoder module with the weights from VACE \citep{vace}, which was trained for video editing using full-attention. We use Low-Rank Adaptation (LoRA) \citep{hu2022lora} to fine-tune the local encoder with rank 128 using the loss $\mathcal{L}$ described in Eq. \ref{eq:total_loss}. We fine-tune a small and a large version of the local encoder consisting of 1.3 billion and 14 billion parameters, respectively. The global embedder module was randomly initialized, with the zero convolution layer's weights and biases set to zero at the start of training. The global token patch layer's weights were copied at the start of training from the video DiT token patch layer. \OurMethod was trained on 8 NVIDIA A100 GPUs for about 1 day with a batch of 8 videos, gradient accumulation of 8, and a learning rate $1e-5$ with AdamW \citep{loshchilov2018adamw} and warmup. Each video in the batch was sampled to 49 frames, and both the frames and masks were down-sampled to $480\times720$. At training and inference, the edit mask is used to set the corresponding area in $\frames_\text{src}$ to $0.5$ to give $\frames_b$.

\paragraph{Evaluation Metrics.}
We evaluate editing and inpainting performance across multiple metrics for the quality of the edit, preservation of regions not meant for editing, and throughput. We measure the masked region preservation (i.e., the background context where no edit is to be done) by comparing it to the groundtruth input and computing the PSNR, SSIM, LPIPS, MSE, and MAE. The semantic alignment between the text prompt and the edited video is measured using the average CLIP score across the full and masked video frames \citep{radford2021clip,hessel2021clipscore}. We follow VideoPainter's \citep{bian2025videopainteranylengthvideoinpainting} procedure for computing temporal coherence and perceptual quality as the average CLIP similarity across all neighboring frames. Finally, we report the inference throughput in terms of frames per second (FPS) on an NVIDIA A6000Ada, excluding VAE encoding and decoding times. We use 25 DDPM iterations at inference for evaluating VACE and \OurMethod.

\subsection{Video Editing \& Inpainting}

To validate \OurMethod's video editing and inpainting approach, we report the average results for the metrics across all videos in the test datasets containing 45 and 150 6-second videos for editing and inpainting, respectively.

\paragraph{Baselines.}
We compare \OurMethod against several state-of-the-art generative and non-generative methods \citep{zhou2023propainter, mou2024revideo,bian2025videopainteranylengthvideoinpainting} as well as VACE \citep{vace} which uses full-attention. The generative methods operate as text-to-video methods except for ReVideo which requires the first frame to be edited. This is provided via an image inpainting backbone \citep{flux2024}.

\paragraph{Results.}
As shown in Tab.~\ref{tab:edit} and Tab.~\ref{tab:inp}, \OurMethod generates edited and inpainted videos that are better aligned with the prompt, better preserve the background, and at a greatly reduced compute load with more than a $50\times$ speedup over generative baselines. Moreover, our trained models exhibit proportional acceleration over the full-attention baseline while matching or even slightly exceeding editing and inpainting quality as well as prompt alignment. Fig.~\ref{fig:editing_visualization} and Fig.~\ref{fig:inpainting_visualization} show qualitative comparison examples of \OurMethod's small model video editing and inpainting results against the baselines, respectively. 
The strong performance of our decoupled framework in both editing and inpainting demonstrates its effectiveness as a general-purpose, high quality and efficient video editing solution.

\begin{figure}[!t]
    \centering
    \includegraphics[width=0.8\linewidth, trim=0.25cm 0pt 0 0pt, clip]{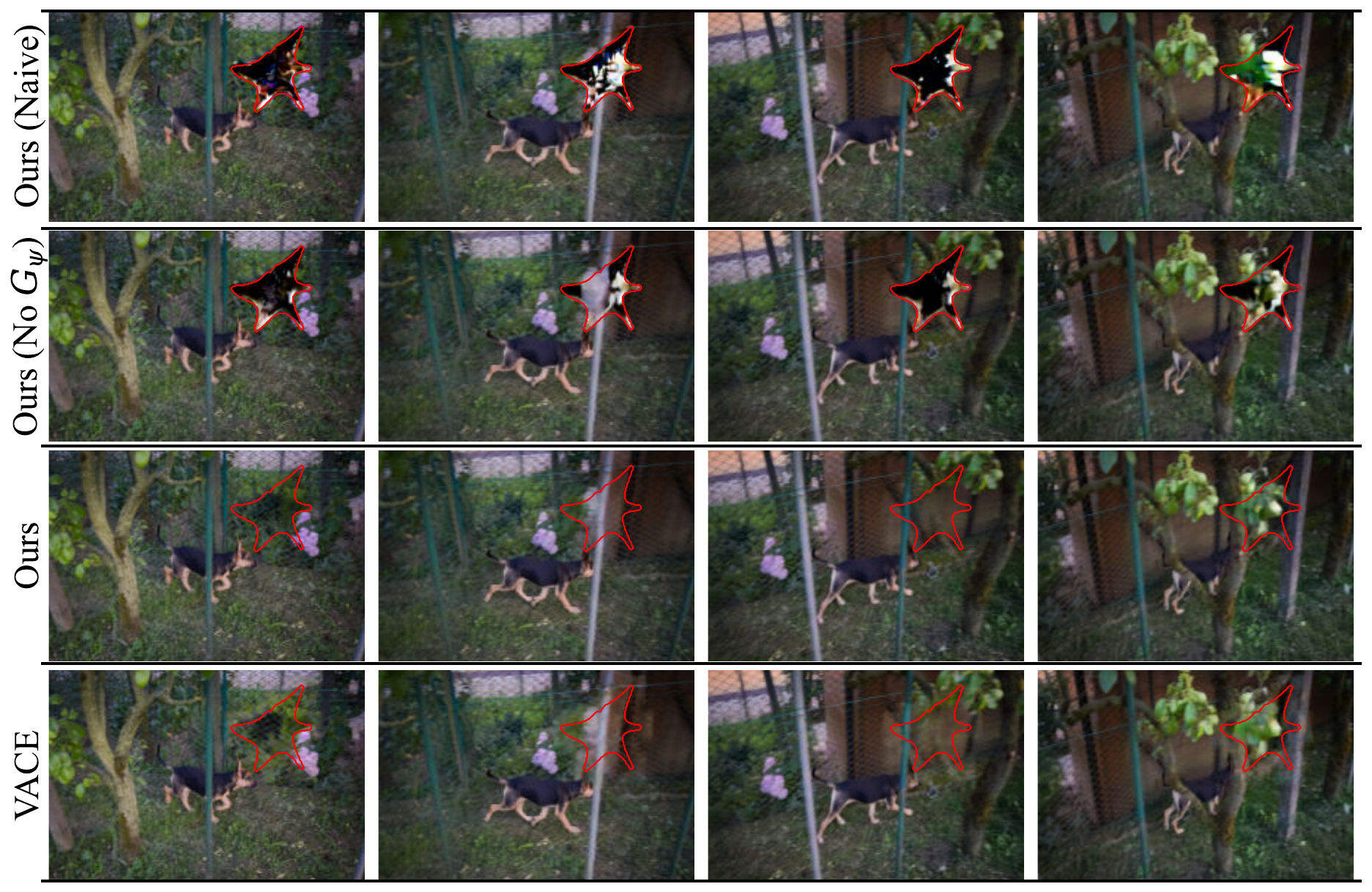}
    \caption{\textbf{Local and Global Adapter Ablation.} Removing adapter components harms video editing quality, but together they let \OurMethod perform comparably to a method operating with full-attention.}
    \label{fig:ablations}
\end{figure}

\begin{table}[t]
\centering
\scriptsize
\setlength{\tabcolsep}{0.6mm}{
\begin{tabular}{clccc|cc|c|c}
\toprule
\multicolumn{2}{c}{} 
& \multicolumn{3}{c}{\bfseries Masked Region Pres.} 
& \multicolumn{2}{c}{\bfseries Text Alignment} 
& \multicolumn{1}{c}{\bfseries Temp. Coherence} 
& \multicolumn{1}{c}{\bfseries Throughput}\\
\cmidrule(lr){3-5} \cmidrule(lr){6-7} \cmidrule(lr){8-8} \cmidrule(lr){9-9} 

 & \bfseries Method 
& PSNR$\uparrow$ & SSIM$\uparrow$ & LPIPS$_{^{\times 10^2}}$$\downarrow$ 
& CLIP$\uparrow$ & CLIP (M)$\uparrow$ & CLIP Sim$\downarrow$  & FPS$\uparrow$ \\
\midrule

& VACE & 23.84 & 0.91 & 5.44 & 9.76 & 21.51 & 0.13 & 0.10 \\ 
& Ours (Naive) & 23.24 & 0.86 & 6.96 & 8.83 & 20.49 & 0.19 & 4.90 \\ 
& Ours (No $G_\psi$) & 23.80 & 0.90 & 5.74 & 9.41 & 21.28 & 0.15 & 4.90 \\ 
& Ours & 24.16 & 0.92 & 5.54 & 9.58 & 21.70 & 0.15 & 4.67 \\ 
\bottomrule
\end{tabular}
}
\caption{\textbf{Quantitative Local and Global Adapter Ablation Comparison.} We observe a clear improvement in quality at a minimal reduction in efficiency with the local and global control modules. With both adapters, \OurMethod greatly increases throughput and even exceeds the full-attention model in edit quality.
}
\label{tab:ablations}
\end{table}

\subsection{Interactive Editing}

While \OurMethod enables fast video editing in target regions, it also lends itself to additional interactive applications such as editing multiple regions with distinct text prompts and real-time content propagation, which we show in Fig.~\ref{fig:teaser}. We show more results for these applications in the Appendix.

\paragraph{Multi-Prompt Editing.} Since generation is performed independently on masked regions, we can easily process multiple, non-contiguous masks simultaneously. Each region's local tokens can even be conditioned on a different text prompt, allowing for complex, multi-prompt edits in a single pass using batch inference.
\paragraph{Content Propagation.} 

Because the base video diffusion model is not fine-tuned for editing, it can be easily swapped out with an autoregressive video diffusion model \citep{huang2025selfforcing} for content propagation and reducing compute even further. By defining the initial edit in one or more frames and propagating the mask, our model can coherently generate the new content across subsequent frames in the local context. Due to the high frame-rate nature of video feeds, the global context does not change much in the near future. Thus, we can treat the global embedding as a causal embedding by padding $\mathbf{V}_b^\downarrow$ with its own last available frames to provide adequate global context about the future. This forgoes the need for global context at inference time and allows us to edit the video into the future. 
The mask can also be propagated forward using motion cues such as optical flow or camera pose, enabling applications such as augmented reality editing where content needs to be generated before the headset acquires the frame and projected to the user once the frame is displayed.

\subsection{Ablation Studies}

We ablate the local encoder and global embedder components from the architecture and compare video editing performance on VPBench-Edit in Tab.~\ref{tab:ablations}. Qualitative comparisons for inpainting on DAVIS are also shown in Fig.~\ref{fig:ablations}. With the naive approach, where we do not use $G_\psi$ and drop all tokens outside the mask before feeding to the non-fine-tuned local encoder, we observe a drastic drop in quality. When using the fine-tuned local encoder without $G_\psi$, editing quality increases drastically but leads to overfitting. Since the local encoder does not know about the global context of the video, it may attend to the prompt too strongly and generate content in the target region, as seen in Fig.~\ref{fig:ablations}. 

\section{Conclusion}
\label{sec:conclusion}

In this work, we introduced \OurMethod, a video diffusion editing approach with disentangled control that generatively edits only where needed. Our method greatly improves over state of the art methods in terms of quality and efficiency and introduces novel interactive functionality, yet limitations remain. First, the video VAE causes a significant degradation to the background context. In addition, the local encoder struggles in videos with very fast motion, which is due to both the VAE and fast shifts in spatiotemporal local context. Lastly, the VAE encode/decode overhead is not a bottleneck for $480\times 720$ videos in terms of end-to-end throughput, but is for 4K videos as we need to tile encode/decode the video due to VRAM constraints. Future work could explore extensions for encoding and integrating additional fundamental temporal information such as motion into generative editing.

\newpage

\bibliographystyle{assets/plainnat}
\bibliography{references}

\newpage

\appendix
\clearpage

\section{Additional Visualizations}
\label{sec:additional_vis}
\begin{figure*}[!t]
    \includegraphics[width=\linewidth, trim=0.05cm 10.25cm 7.55cm 0, clip]{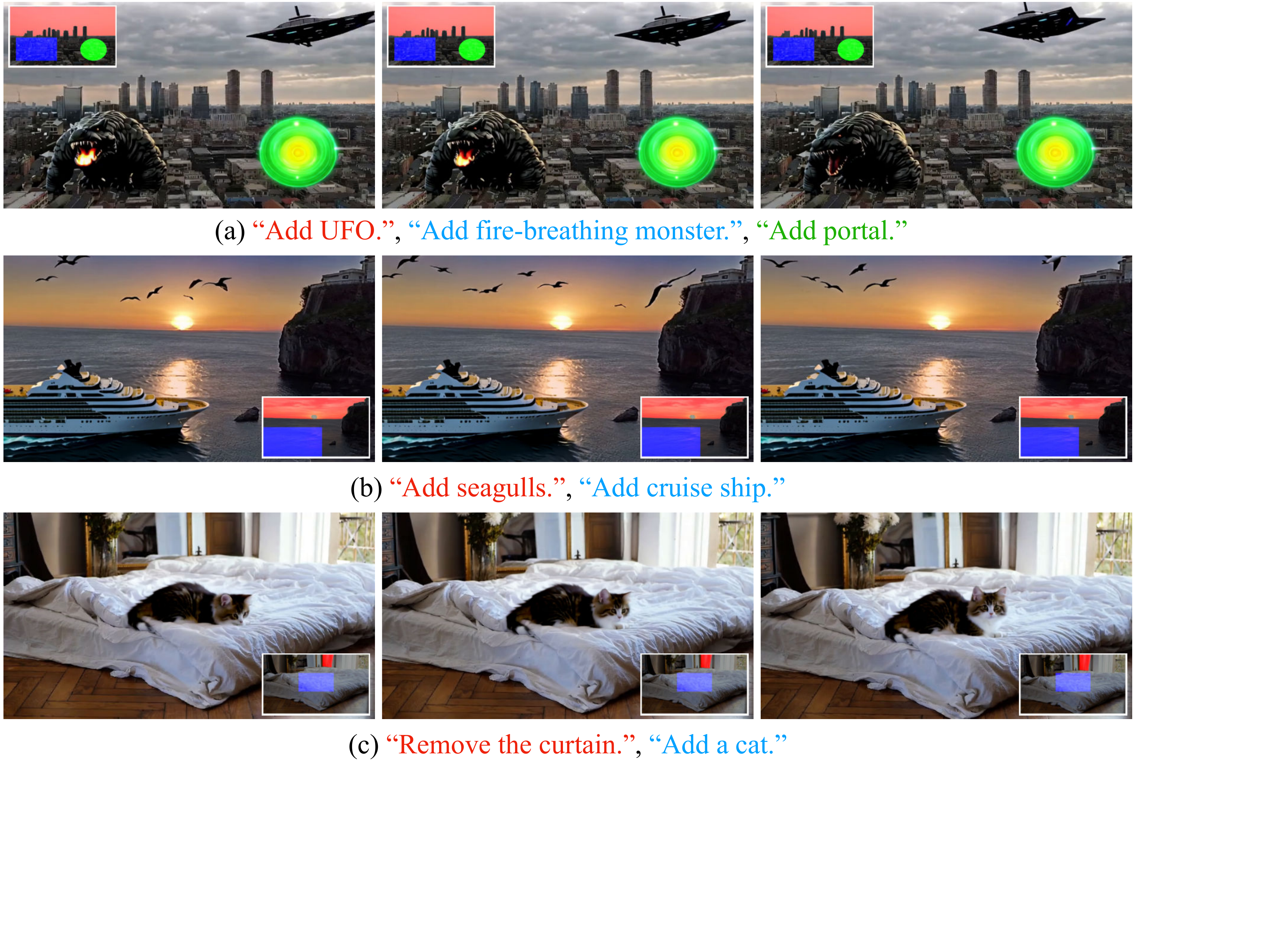}
    \caption{\textbf{Additional Multi-Prompt Generation Visualizations.}}
    \label{fig:additional_mp_visualizations}
\end{figure*}

\begin{figure*}[!t]
    \includegraphics[width=\linewidth, trim=0.05cm 0 7.55cm 0, clip]{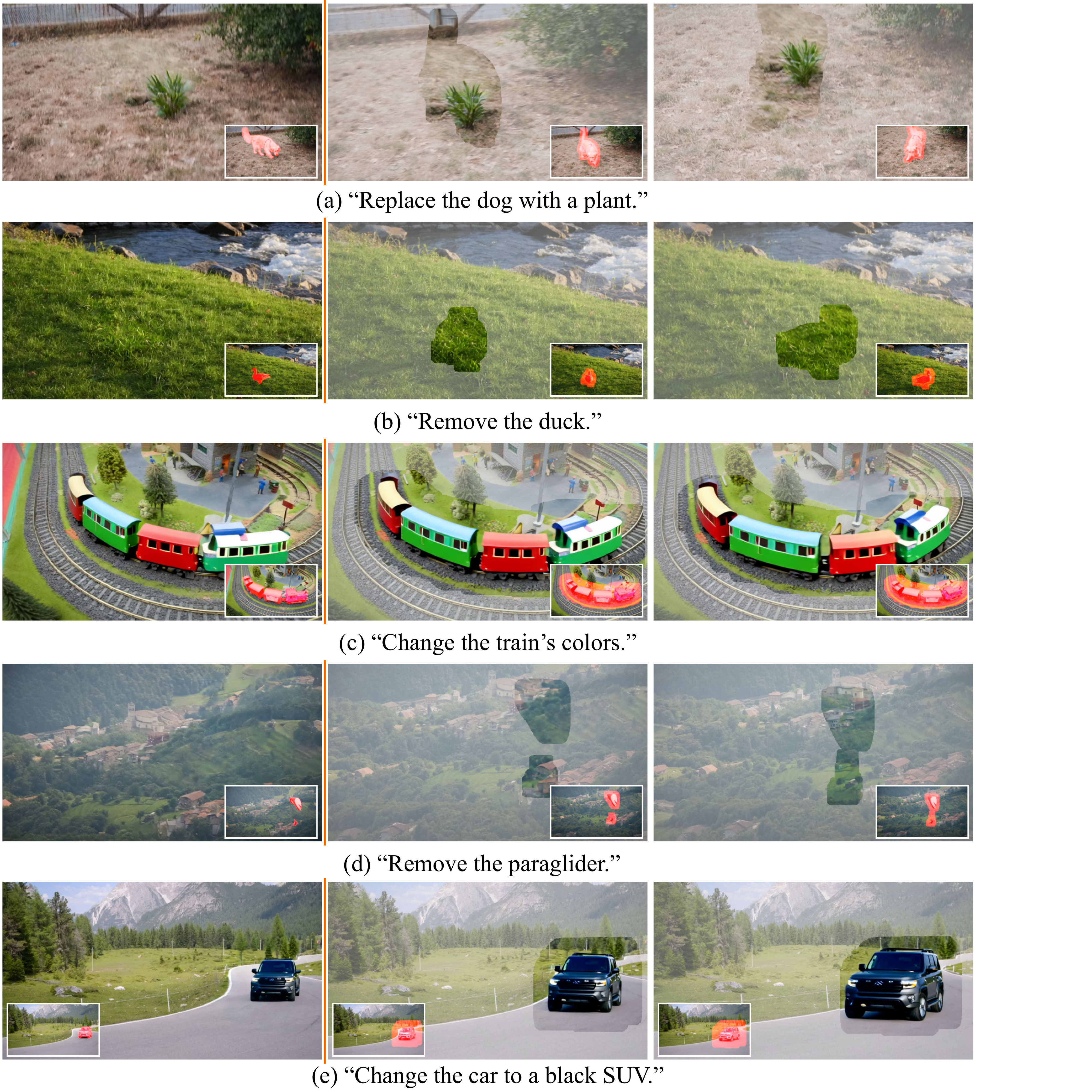}
    \caption{\textbf{Additional Content Propagation Visualizations.}}
    \label{fig:additional_propagation_visualizations}
\end{figure*}

\begin{figure*}
    \centering
    \captionsetup{type=figure}
        \includegraphics[width=0.778\textwidth, trim=0.26cm 0pt 0 0pt, clip]{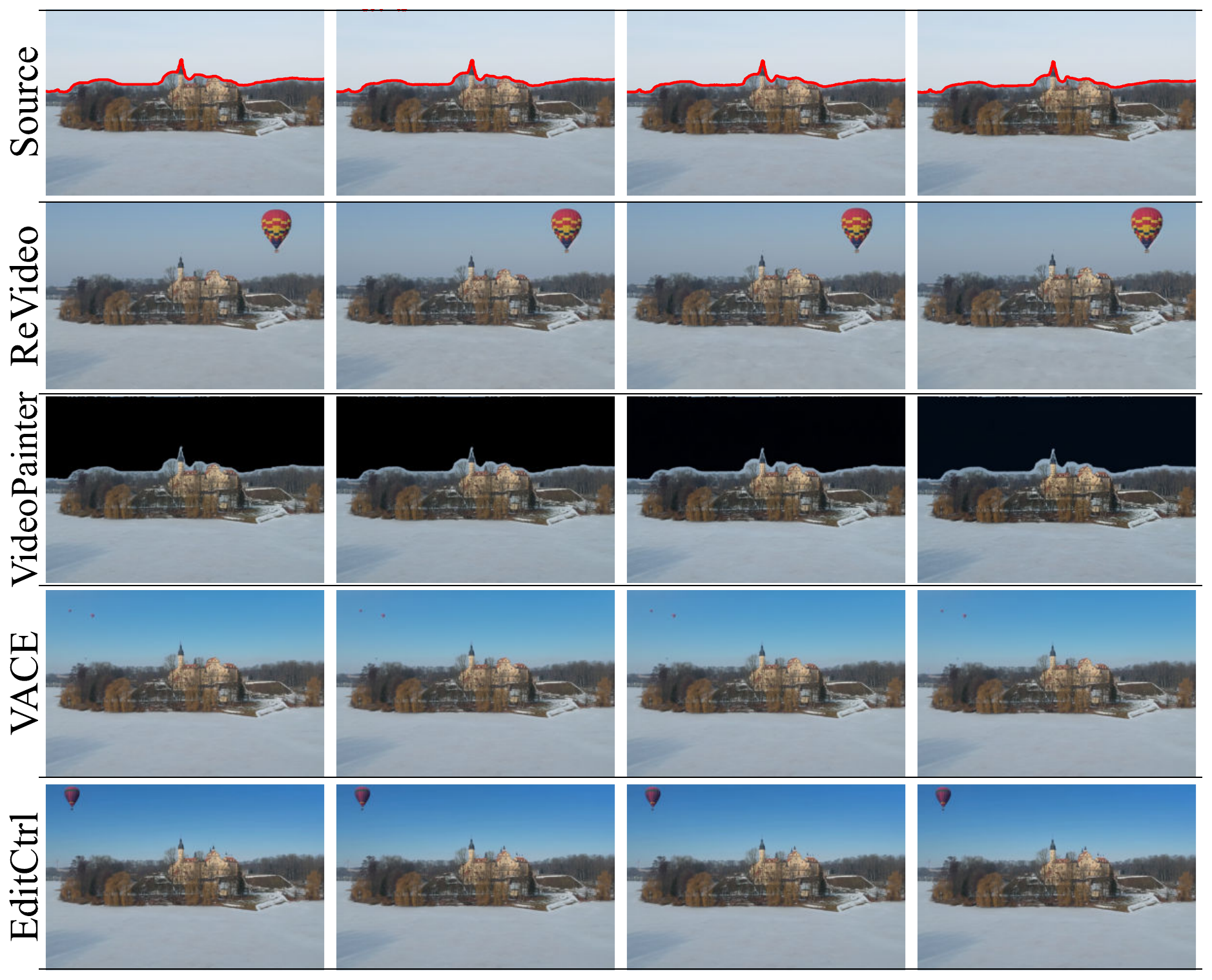}
        \hspace*{-0.2cm}
        \includegraphics[width=0.778\textwidth, trim=0pt 0pt 0.26cm 0pt, clip]{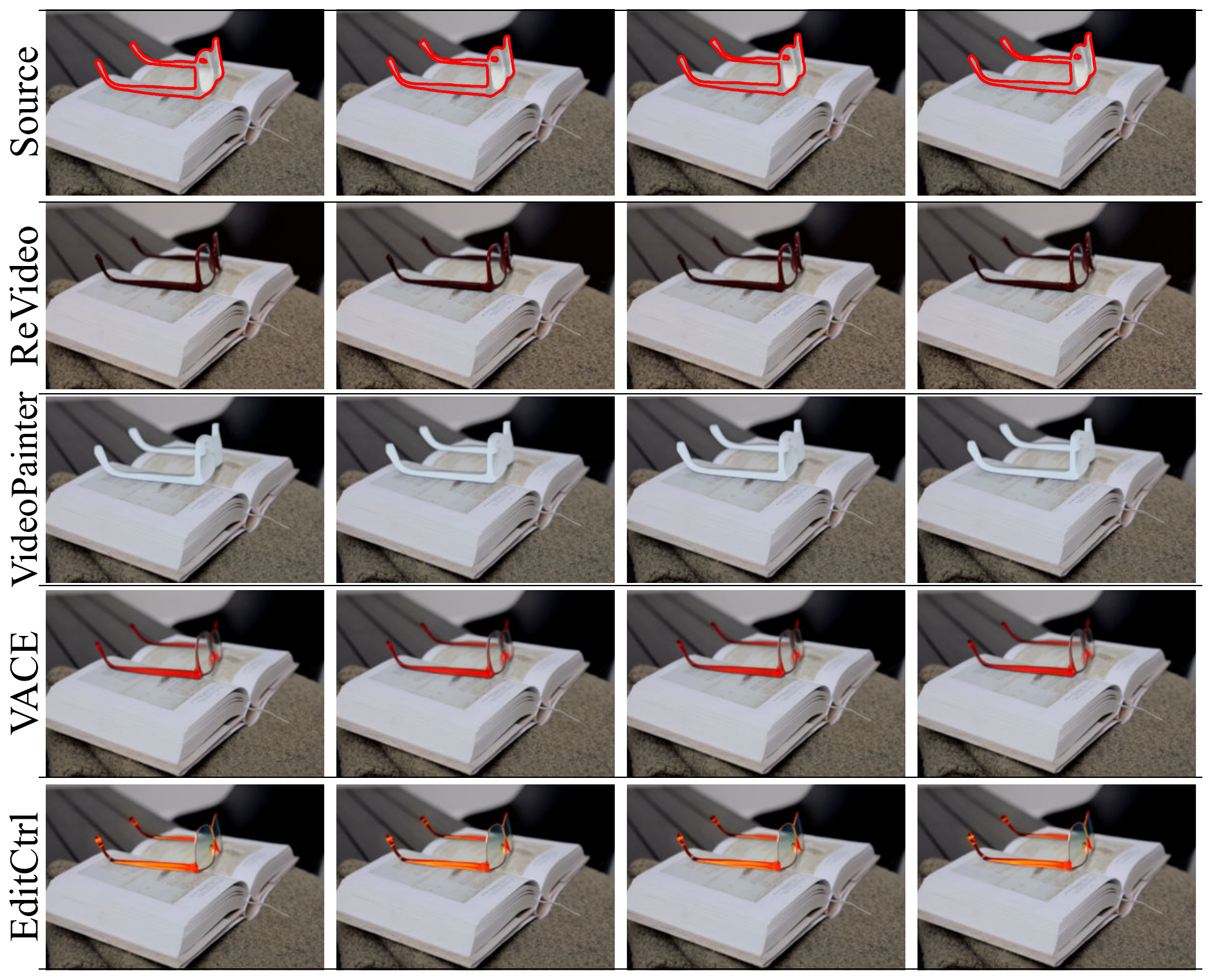}
     \vspace*{-0.4cm}
       \captionof{figure}{
            \textbf{Additional Video Editing Comparisons.} (Top) "Add a hot air balloon" (Bottom) "Turn the sunglasses shiny red."
        }
        \label{fig:additional_editing_visualizations_1}
\end{figure*}

\begin{figure*}
    \centering
    \captionsetup{type=figure}
        \includegraphics[width=0.778\textwidth, trim=0.26cm 0pt 0 0pt, clip]{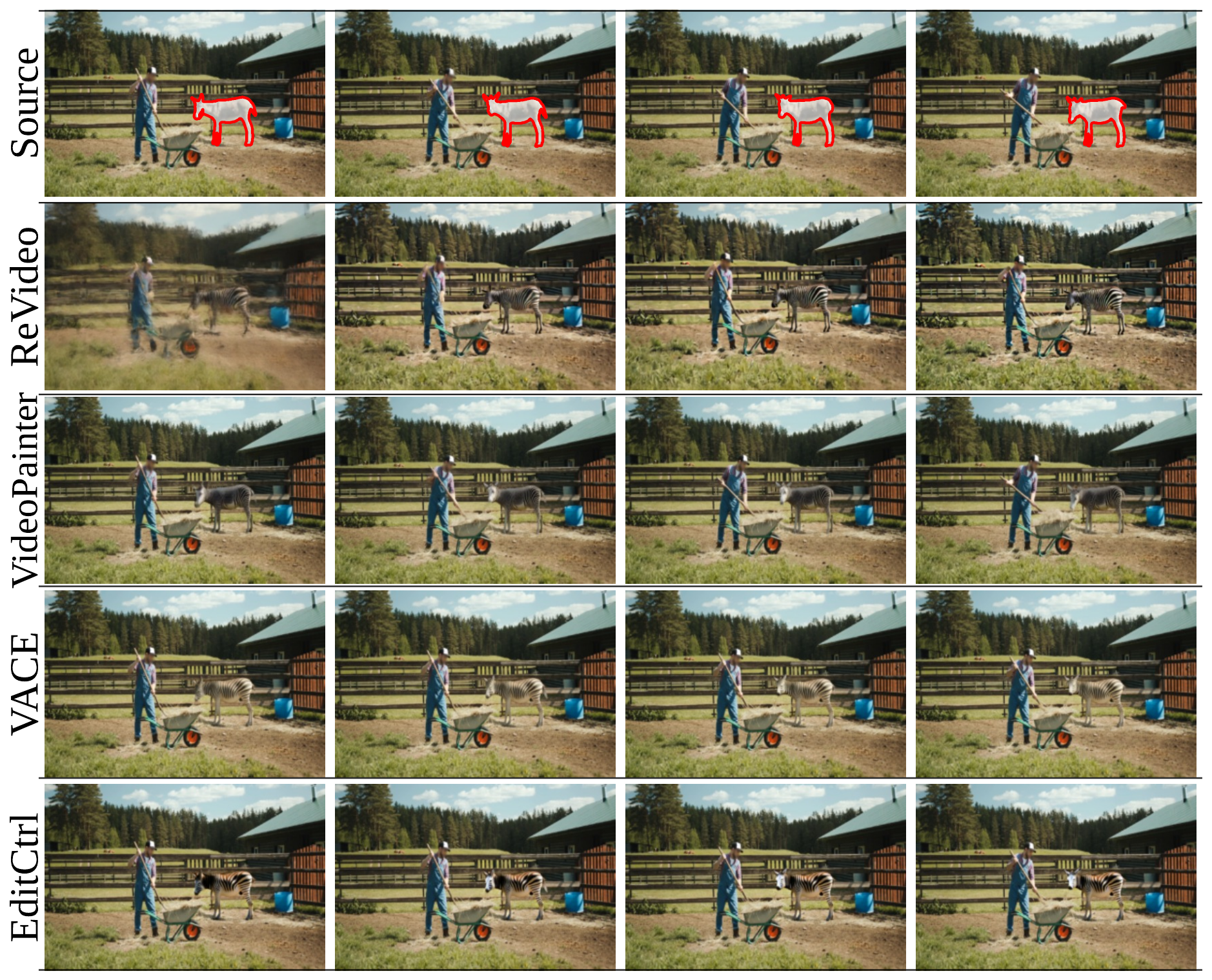}
        \hspace*{-0.2cm}
        \includegraphics[width=0.778\textwidth, trim=0pt 0pt 0.26cm 0pt, clip]{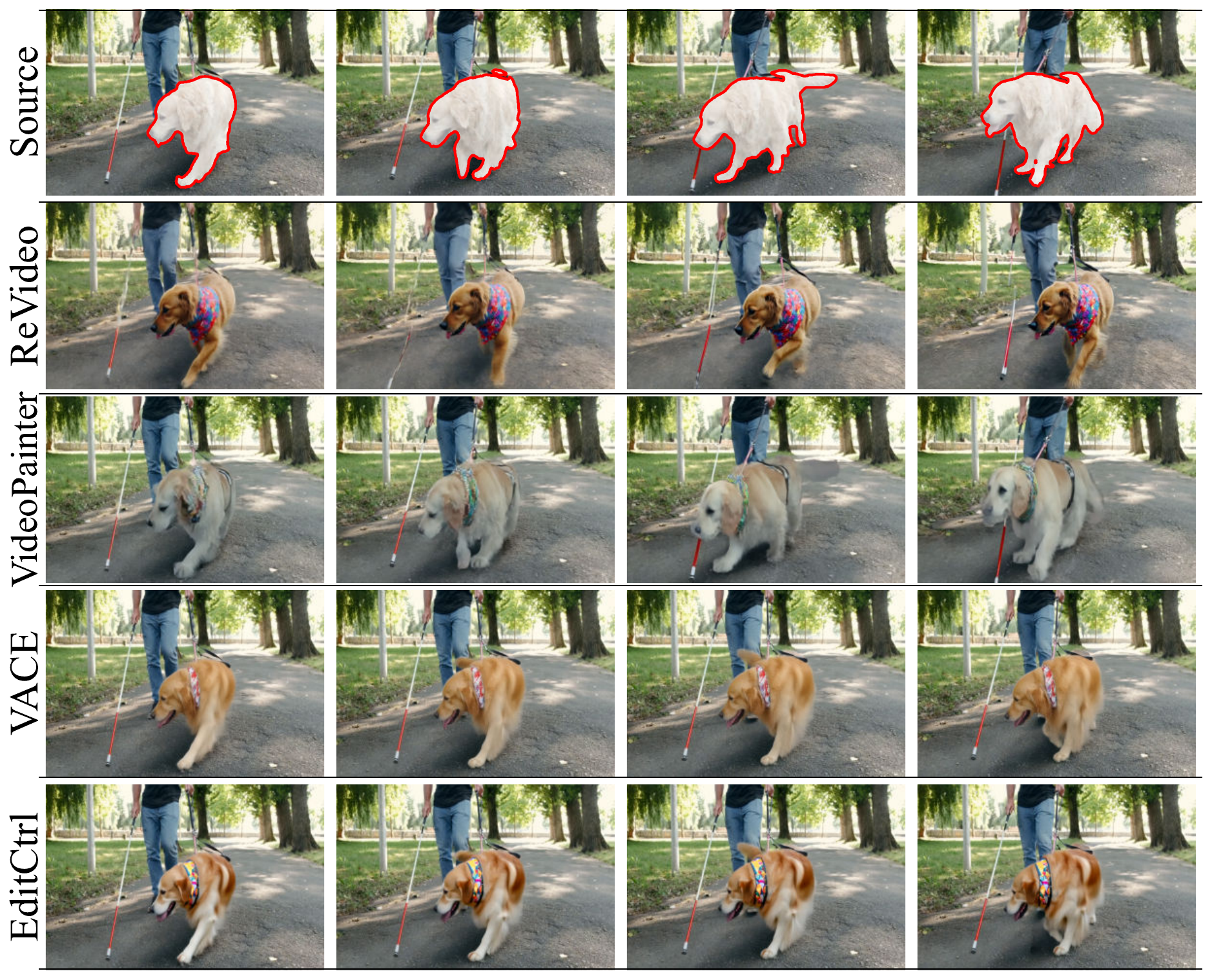}
     \vspace*{-0.4cm}
       \captionof{figure}{
            \textbf{Additional Video Editing Comparisons.} (Top) "Turn the donkey into a zebra." (Bottom) "Add a bandana to the dog."
        }
        \label{fig:additional_editing_visualizations_2}
\end{figure*}

\begin{figure*}
    \centering
    \captionsetup{type=figure}
        \includegraphics[width=0.778\textwidth, trim=0.26cm 0pt 0 0pt, clip]{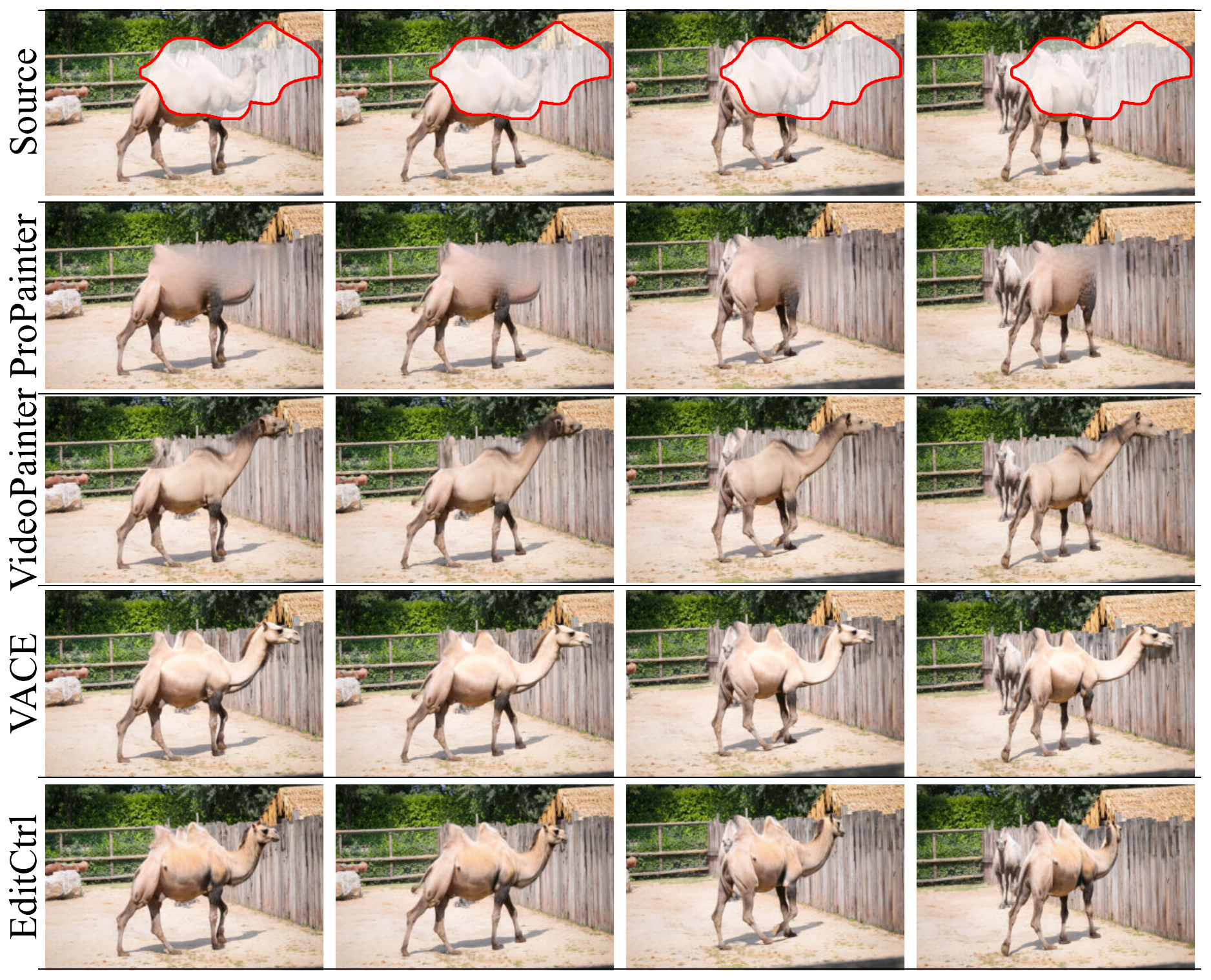}
        \hspace*{-0.2cm}
        \includegraphics[width=0.778\textwidth, trim=0pt 0pt 0.26cm 0pt, clip]{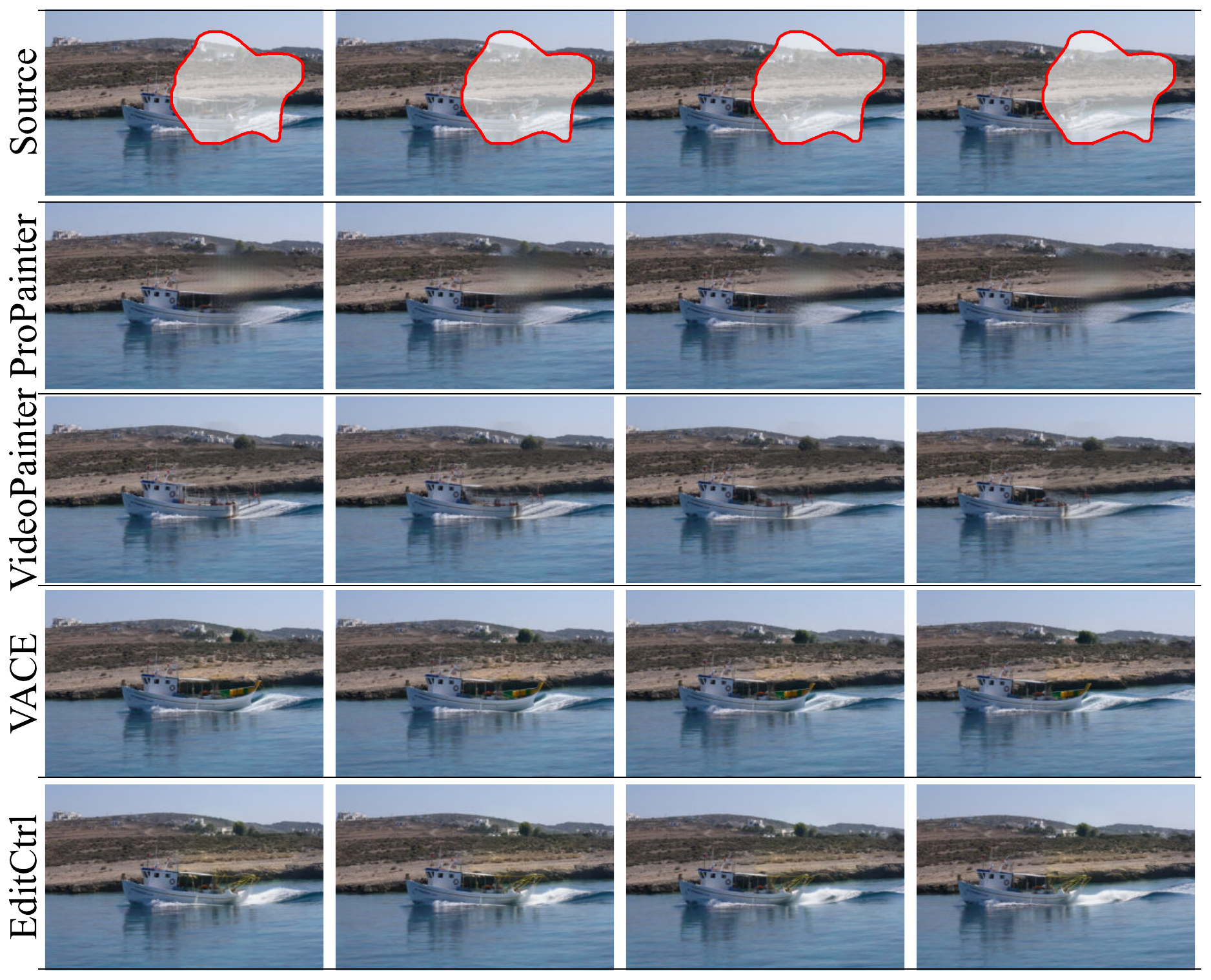}
     \vspace*{-0.4cm}
       \captionof{figure}{
            \textbf{Additional Video Inpainting Comparisons on DAVIS.}
        }
        \label{fig:additional_inpainting_visualizations_davis}
\end{figure*}

\begin{figure*}
    \centering
    \captionsetup{type=figure}
        \includegraphics[width=0.778\textwidth, trim=0.26cm 0pt 0 0pt, clip]{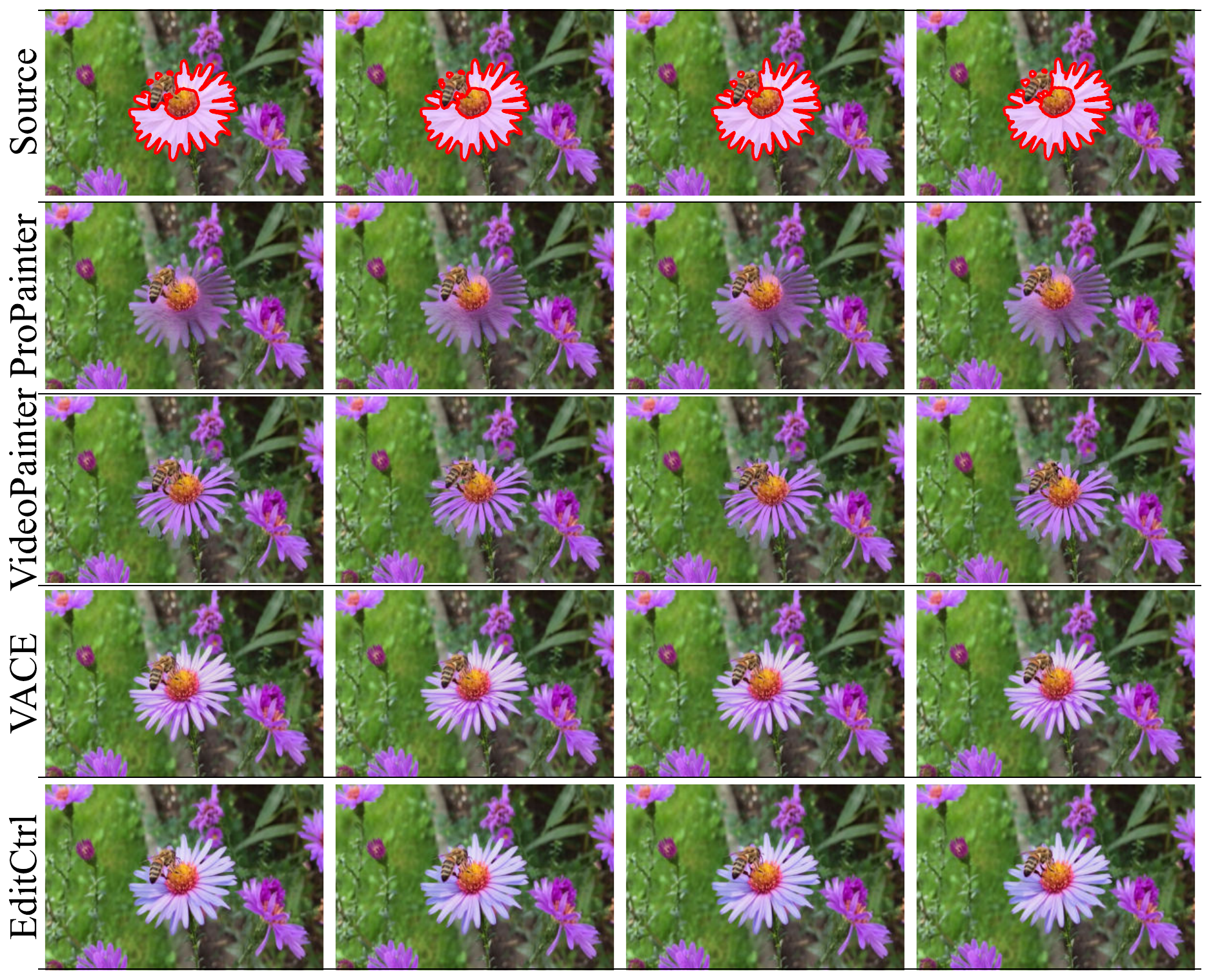}
        \hspace*{-0.2cm}
        \includegraphics[width=0.778\textwidth, trim=0pt 0pt 0.26cm 0pt, clip]{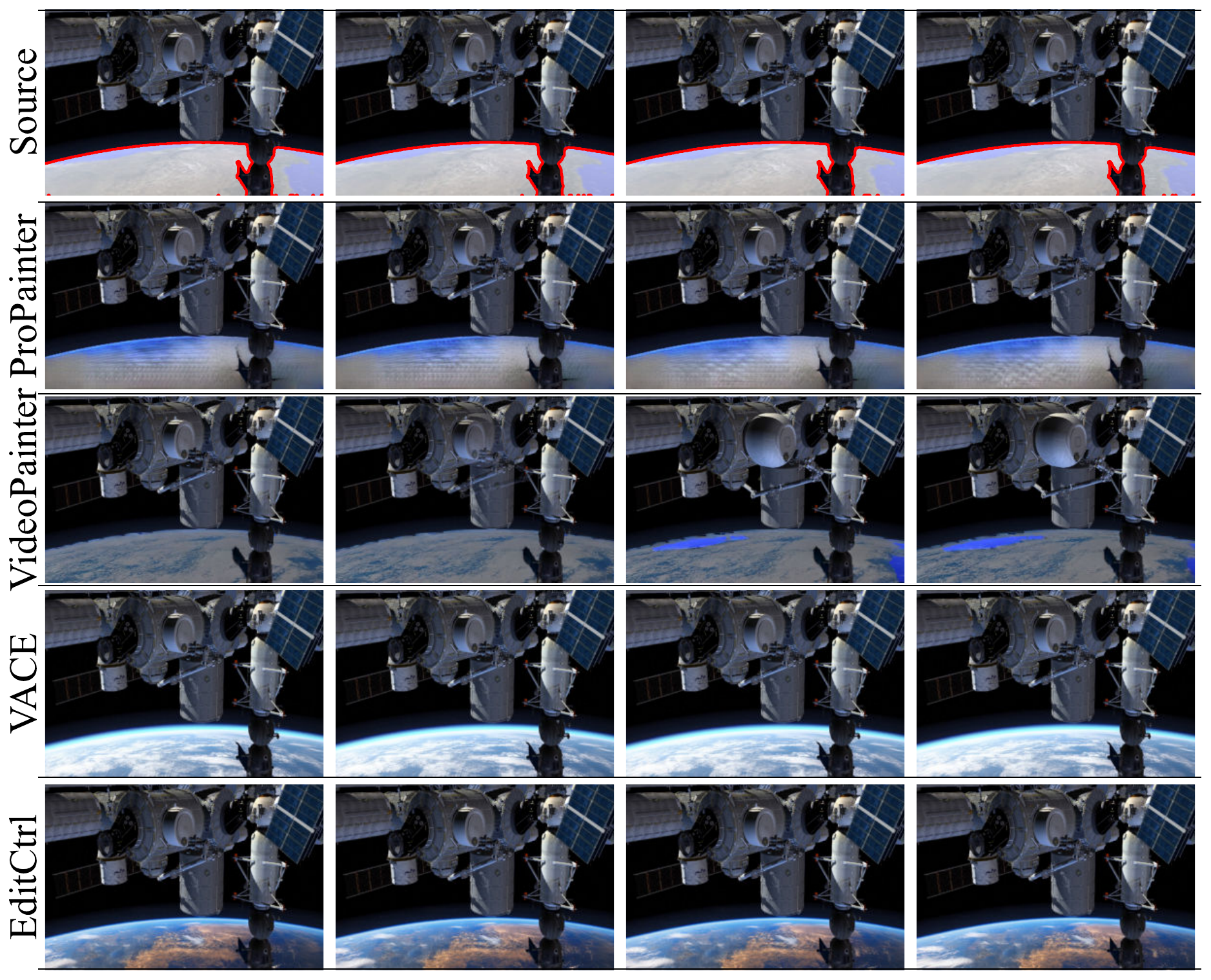}
     \vspace*{-0.4cm}
       \captionof{figure}{
            \textbf{Additional Video Inpainting Comparisons on VPBench.}
        }
        \label{fig:additional_inpainting_visualizations_vpbench}
\end{figure*}

We show additional results editing in distinct regions with multiple prompts as well as content propagation in Figs.~\ref{fig:additional_mp_visualizations}-\ref{fig:additional_propagation_visualizations}. Furthermore, we provide additional visualizations of \OurMethod's video editing and inpainting results on VPBench and DAVIS in Figs.~\ref{fig:additional_editing_visualizations_1}-\ref{fig:additional_inpainting_visualizations_vpbench}.

\section{Additional Details}
\label{sec:additional_details}

\begin{figure}[!h]
\vspace*{-3mm}
    \centering
    \includegraphics[width=0.36\linewidth, trim=0.05cm 5.64cm 30.78cm 0, clip]{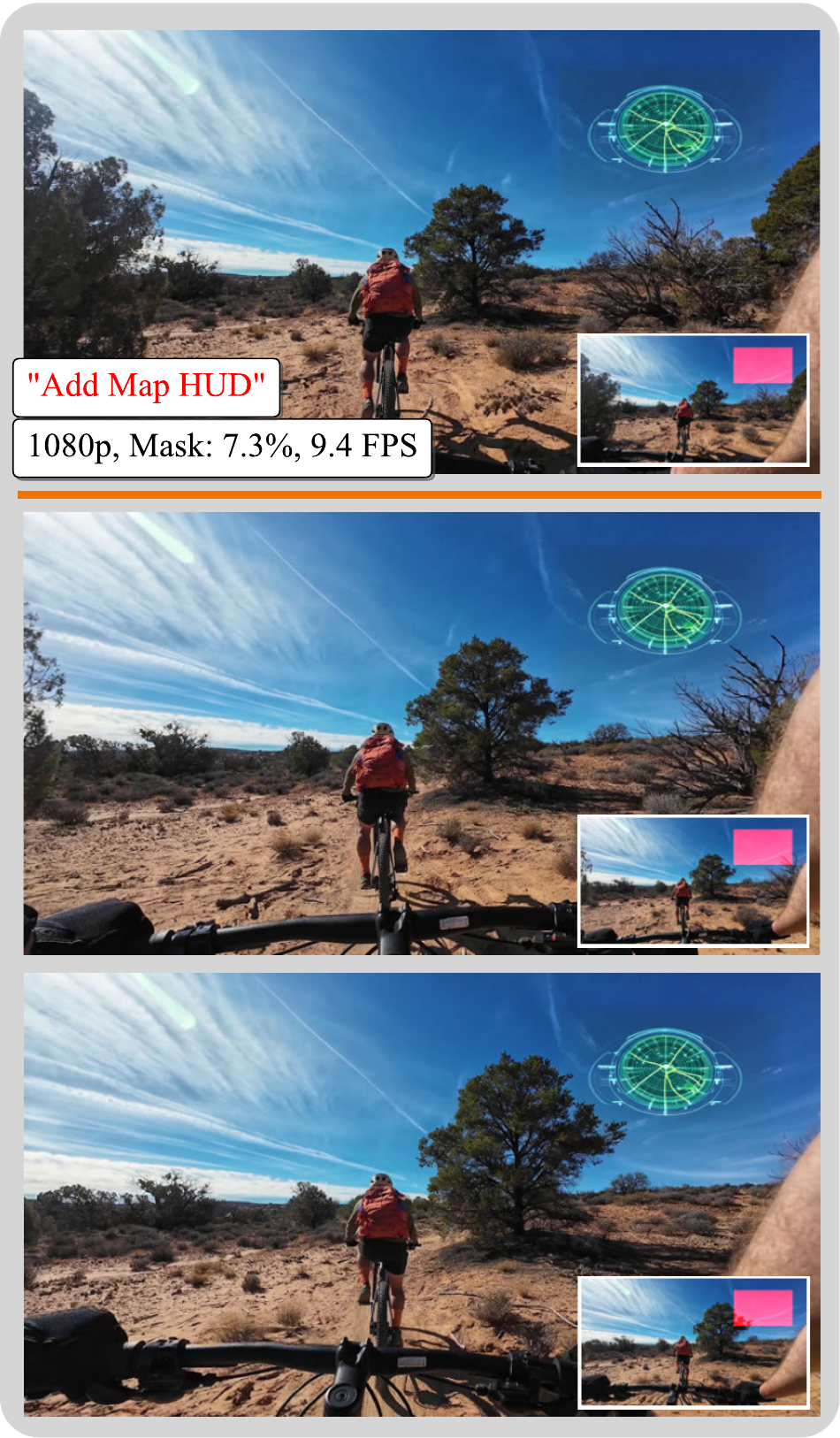}
    \caption{\textbf{Content Propagation for Augmented Reality.} \OurMethod is particularly suitable for deployment in augmented reality applications given its low latency and ability to propagate content to match the user's movement.}
    \label{fig:ar_example}
\end{figure}

\paragraph{Content Propagation.}
When using a distilled autoregressive video diffusion model, video generation can be extended to an indefinite range, albeit with some loss in quality by denoising new incoming noisy latents. When used as the base video model for \OurMethod, this means we inherit the autoregressive model's properties and can edit a video of any length by using sliding window attention over a local window of frames. This is in addition to accelerated inference with respect to the target edit area. Moreover, we can extend this to real-time video editing, where we edit a video indefinitely, but without having access to the background context. This is achieved by simply copying the background context from the last known frames. If the video is static then the background context tokens used for local context can be copied over for generating next frames. In cases where the background context is dynamic, we use optical flow to propagate the last frame's pixels to roughly approximate the future local context. In both static and dynamic cases, the last downsampled background frame $\mathbf{V}_{b_N}^\downarrow$ is padded to $\mathbf{V}_{b}^\downarrow$ to provide a rough estimate for future global context. Once the future frame arrives, the generated content is pasted into the corresponding pixels along with optional edge blending. This process is repeated, where optical flow propagation is computed from the last known frame to the next future frame. These steps help generate more visually appealing content that matches the input video's motion ahead of time, whereas if we wait for future frames the local context will be much more accurate but the latency requirement would be too small, making the generated output appear laggy. An example of \OurMethod in an augmented reality setting is shown in Fig.~\ref{fig:ar_example}, where initial frames are known and then \OurMethod propagates content to the next frames.

\begin{figure*}[!t]
    \centering
    \includegraphics[width=\linewidth, trim=0.05cm 0.56cm 0pt 0, clip]{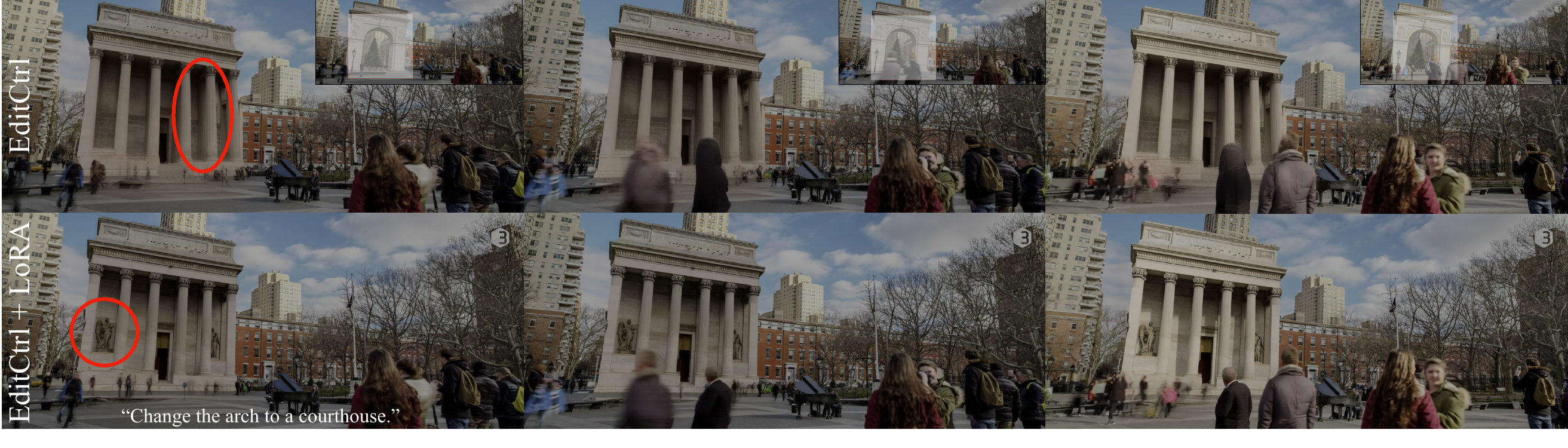}
    \caption{\textbf{Editing complex timelapse 21:9 ultrawide footage.} EditCtrl can also be used with additional conditioning LoRAs \citep{diffsynth_studio_wan2.1_1.3b_lora_aesthetics_v1}.}
    \label{fig:style_transfer}
\end{figure*}

\paragraph{$G_\psi$ Resolution \& Transferability.} 
To test the fixed 256$\times$256 global embedder resolution and transferability, we test EditCtrl on a challenging 21:9 ultrawide timelapse video by itself and with a publicly available aesthetic style LoRA \citep{diffsynth_studio_wan2.1_1.3b_lora_aesthetics_v1} in Fig.~\ref{fig:style_transfer}. This example demonstrates that the downsampled global context captures sufficient spatial and temporal information even for extreme aspect ratios and complex scenes, and moreover, that our adapter design composes seamlessly with style LoRAs, confirming the non-destructive design generalizes beyond the base model. This is shown by the enhanced style EditCtrl generates when using an aesthetic LoRA.

\begin{figure*}[!t]
    \centering
    \includegraphics[width=\linewidth]{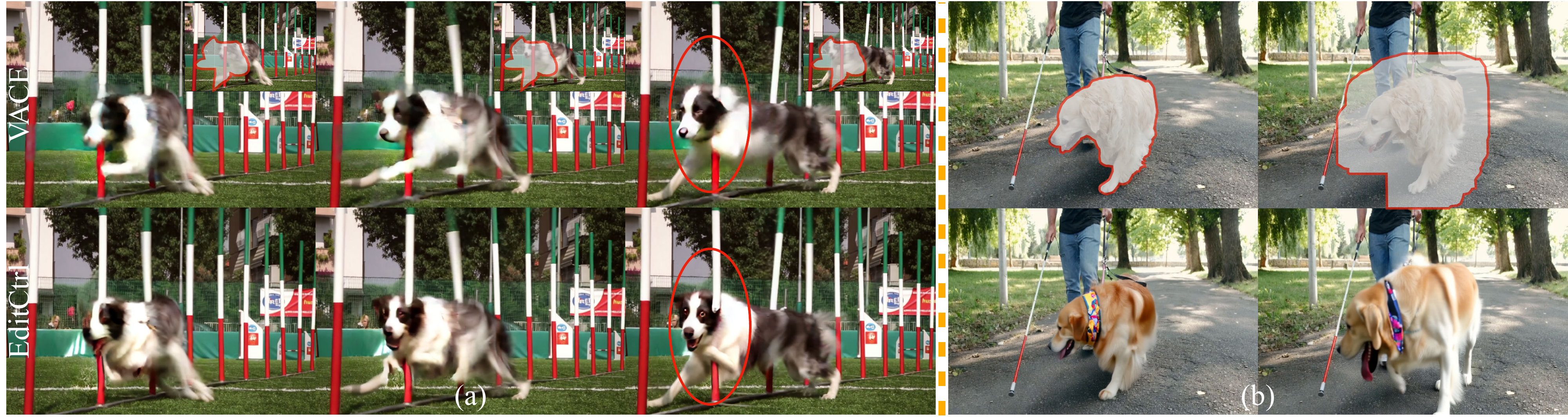}
    \caption{\textbf{Failure Modes for \OurMethod on High Motion (a) \& Inaccurate Masks (b).}}
    \label{fig:failure_modes}
\end{figure*}

\paragraph{Failure Modes.} 
We present two cases of inputs where EditCtrl fails in Fig.~\ref{fig:failure_modes}. In the first case on the left, at high motion EditCtrl introduces artifacts and incorrect interaction with the surrounding scene. The artifacts are also present for the full-attention baseline VACE, indicating that this is a structural problem that we attribute to the VAE. In the second case on the right, as the mask becomes exceedingly large finegrained editing becomes more difficult and EditCtrl changes the content by more than desirable.

\newpage

\newpage


\end{document}